\documentclass{article}

\usepackage[preprint]{neurips_2026}

\usepackage[utf8]{inputenc}
\usepackage[T1]{fontenc}
\usepackage[hidelinks]{hyperref}
\usepackage{url}
\usepackage{amsmath,amssymb,amsfonts}
\usepackage{graphicx}
\usepackage{xcolor}
\usepackage{nicefrac}
\usepackage{microtype}
\usepackage{longtable,booktabs,array}
\usepackage{calc}
\usepackage{etoolbox}
\usepackage{wrapfig}
\makeatletter
\patchcmd\longtable{\par}{\if@noskipsec\mbox{}\fi\par}{}{}
\makeatother
\usepackage{footnote}
\makesavenoteenv{longtable}
\providecommand{\tightlist}{%
  \setlength{\itemsep}{0pt}\setlength{\parskip}{0pt}}
\usepackage{bookmark}

\title{Synthetic Benchmarks Overstate Forward-Forward Scaling:\\Real-Data Limits of Layer-Local Training}

\author{
  Yucheng Chen\thanks{Corresponding author.} \\
  Amplimit \\
  Toronto, ON, Canada \\
  \texttt{steven.chen@amplimit.com} \\
}

\begin{document}

\maketitle

\begin{abstract}
	Forward-Forward (FF) learning \citep{hinton2022forward} replaces backpropagation with strictly layer-local goodness updates. Recent FF-CNN work has narrowed the gap to BP on $32\!\times\!32$ benchmarks, raising the question of whether layer-local training is becoming a viable alternative at realistic scale. To probe this rigorously, we develop \textbf{DTG-FF}---dynamic temperature goodness, decoupled normalization, and multi-layer fusion---as an instrument that sets FF-family state of the art across nine real-data benchmarks at the time of our submission ($91.8\%$ CIFAR-10 and an FF baseline at ImageNet-100 $224\!\times\!224$), and use it to audit how far layer-local training actually scales.
	\textbf{(1)~Real-data scaling.} Under identical recipe and backbone, an architecture-matched BP-DeepSup baseline beats DTG-FF by $2.40$/$5.93$ pp on CIFAR-10/CIFAR-100, and the gap widens with class count. At $224\!\times\!224$ the same instrument reaches only $49.4\%$ on ImageNet-100, versus $75.6\%$ even for a self-supervised BP-trained ResNet-50 with linear evaluation \citep{tian2020cmc}---exposing a real-data ceiling invisible at $32\!\times\!32$.
	\textbf{(2)~Synthetic vs.\ real $K$-conflict.} DTG-FF increasingly outperforms BP as class count $K$ grows on synthetic teacher-student tasks, yet on real images the FF--BP gap reverses sign and widens with $K$. A within-dataset CIFAR-100 coarse vs.\ fine probe isolates label-hierarchy from image distribution: synthetic $K$-sweeps confound output dimensionality with fine-grained discrimination difficulty and thereby overstate FF transferability.
	\textbf{(3)~Systems audit.} FF can be implemented without storing depth-wide activations, but on commodity 8\,GB hardware standard BP+gradient-accumulation reaches $4.18$\,GB / $157$ imgs/s versus DTG-FF's $7.90$\,GB / $138$ imgs/s, so a memory-based justification for FF at this scale is not supported under fair baselines.
\end{abstract}

\section{Introduction}\label{introduction}

Backpropagation's global backward pass couples all layers and stores activations across depth, motivating strictly layer-local training rules as potential accuracy--memory tradeoffs rather than direct BP replacements. The Forward-Forward (FF) algorithm \citep{hinton2022forward} is one such rule: it replaces the backward pass with layer-local goodness-based learning, training each layer to produce high ``goodness'' (squared activation norm) for positive data and low goodness for negative data, with no cross-layer gradient flow. FF is sometimes also motivated by biological considerations \citep{crick1989,lillicrap2020backprop}, but our interest here is empirical: is layer-local training scalable enough to be a useful alternative on real-data workloads, and what systems property does it actually deliver?

FF has not yet demonstrated this competitiveness. On CIFAR-10 the original FF algorithm reaches roughly $59\%$ with locally connected, non-convolutional networks \citep{hinton2022forward}, about $35$ points below convolutional BP. Subsequent work has narrowed the gap through architectural innovations---convolutional extensions \citep{papachristodoulou2024cwcomp,chen2025scff}, deeper architectures \citep{sun2025deeperforward}, and adaptive spatial goodness encoding \citep{gong2026asge}---reaching $90.62\%$ (ASGE VGG11) but still trailing BP baselines, and almost exclusively on $32\!\times\!32$ inputs.

We test this with a strong FF-family instrument and make four contributions:

\textbf{1.~An instrument for stress-testing layer-local training.}
We develop DTG-FF as a compact FF-family architecture combining three mechanism-level improvements (dynamic temperature goodness, decoupled three-path normalization, multi-layer fusion). At the time of our submission (May 2026) it set FF-family state of the art on nine real-world benchmarks (Sec.~\ref{sec:sota}; concurrent work has since reported higher CIFAR-100 and Tiny ImageNet results)---CIFAR-10 ($91.79\%$ logit-sum / $91.33\%$ concat), CIFAR-100 ($67.28\%$), Tiny ImageNet ($48.17\%$), and an FF-family ImageNet-100 baseline at $224\!\times\!224$ ($49.4\%$ on VGG11; ASGE \citep{gong2026asge} previously reported FF training on ImageNet-1k). We treat this not as the paper's headline claim but as the credibility floor for the audit that follows: only a strong FF instrument can tell us whether the residual FF--BP gap reflects fundamental limits of layer-local training or merely the weakness of prior FF baselines.

\textbf{2.~Real-data scaling diagnosis with architecture-matched BP controls.}
Despite FF-family SOTA, DTG-FF (concat) trails an architecture-matched BP-DeepSup baseline under identical recipe and backbone by $2.40$ pp on CIFAR-10 ($K\!=\!10$) and $5.93$ pp on CIFAR-100 ($K\!=\!100$); the FF--BP gap widens with class count. A deeper-and-narrower VGG11 backbone underperforms VGG8 on $32\!\times\!32$ inputs by $-6.97$/$-12.49$ pp on CIFAR-10/100 (App.~\ref{app:depth}), but VGG11 differs from VGG8 in both depth (8 vs.\ 7 conv layers) and early-channel width (64 vs.\ 128 starting channels), so we cannot strictly attribute the drop to depth alone. At $224\!\times\!224$ the same instrument reaches only $49.4\%$ on ImageNet-100 versus $75.6\%$ even for a self-supervised BP-trained ResNet-50 with linear evaluation~\citep{tian2020cmc}, exposing a real-data ceiling invisible at $32\!\times\!32$.

\textbf{3.~Synthetic--real $K$-axis conflict.}
In paired-seed teacher--student synthetic tasks DTG-FF's advantage over BP \emph{grows} with $K$; on real images the FF--BP gap reverses sign and \emph{widens} with $K$. A within-dataset CIFAR-100 coarse-$20$ vs.\ fine-$100$ probe (Sec.~\ref{sec:granularity}) shows that $K$ in synthetic sweeps tracks output dimensionality whereas $K$ on real data also tracks fine-grained discrimination difficulty; current synthetic FF validation overstates real-data transferability.

\textbf{4.~A fair-baseline systems audit.}
A natural defense of FF is its $\mathcal{O}(1)$-in-depth activation-memory property. Pipelined per-layer training realizes this bound, but on commodity 8\,GB hardware standard BP+gradient-accumulation reaches $4.18$\,GB / $157$ imgs/s versus DTG-FF's $7.90$\,GB / $138$ imgs/s, so a memory-based justification for FF at this scale is not supported (Sec.~\ref{sec:systems}, App.~\ref{app:memory}). We additionally offer an interpretive synthesis (Sec.~\ref{sec:bp_shadow}) reading several FF-family improvements---label overlay, BP-trained classifier heads, spatial goodness, multi-layer fusion---as partial substitutes for the supervised cross-layer signal that BP provides natively.

\begin{figure}[t]
	\centering
	\includegraphics[width=\textwidth]{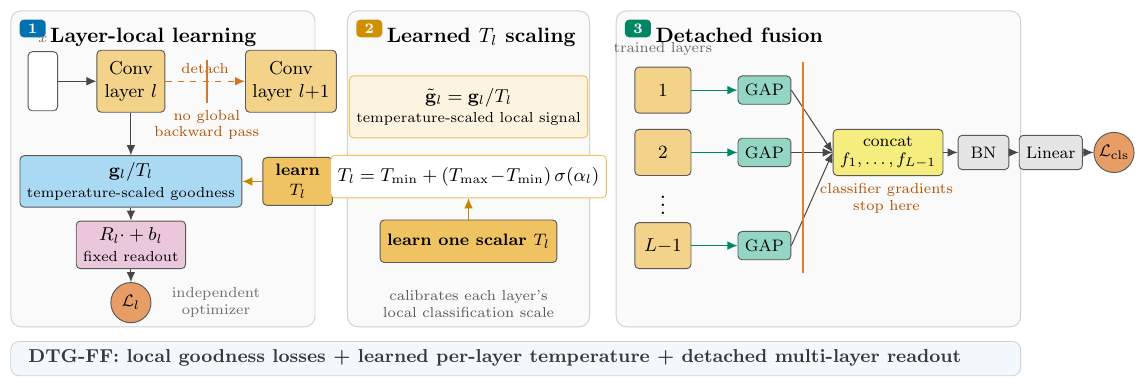}
	\caption{\textbf{DTG-FF method overview.} The method combines three mechanisms: layer-local FF losses with detached propagation, a learnable per-layer temperature $T_l$ that scales spatial goodness before the fixed random readout, and a detached multi-layer classifier that fuses GAP features through BN$+$Linear without updating the convolutional backbone.}
	\label{fig:method}
\end{figure}

Section~\ref{sec:related} surveys related work. Section~\ref{sec:analysis} presents the per-layer signal diagnostic and the synthetic validation with architecture-matched BP controls. Section~\ref{sec:method} describes DTG-FF. Section~\ref{sec:experiments} reports real-data scaling and the within-dataset $K$-disambiguation. Section~\ref{sec:discussion} reports the fair-baseline systems audit and offers an interpretive reading of FF-family improvements as partial substitutes for BP's cross-layer supervised signal.

\section{Related Work}\label{sec:related}

\textbf{Forward-Forward and variants.}
FF was introduced by \citet{hinton2022forward} with each layer trained to produce high goodness on positive and low on negative data. Subsequent work has reduced the FF--BP accuracy gap: CwComp \citep{papachristodoulou2024cwcomp} introduced channel-wise competitive learning for FF-CNNs (78.11\% CIFAR-10); SCFF \citep{chen2025scff} builds positive and negative examples self-contrastively and trains without labels before a linear readout (80.75\%); DeeperForward \citep{sun2025deeperforward} combines layer normalization with mean goodness to train deeper FF networks (88.72\% with 1000 epochs and stronger augmentation; 86.22\% in its main protocol); ASGE \citep{gong2026asge} supervises spatial goodness encodings layer-wise and fuses concatenated features (90.62\% on VGG11, the prior FF-family best at the time of our submission). \citet{ororbia2023pff} combined FF with predictive coding. Outside FF, SoftHebb \citep{journe2023softhebb} reaches 80.31\% with unsupervised soft winner-take-all Hebbian learning and a linear readout.

\textbf{Other biologically motivated alternatives to BP.}
Feedback Alignment \citep{lillicrap2016feedback,nokland2016direct,launay2020dfa}, Target Propagation \citep{bengio2014target,lee2015target,meulemans2020theoretical}, Equilibrium Propagation \citep{scellier2017equilibrium,laborieux2024improving,millidge2023backpropagation}, and Predictive Coding \citep{rao1999predictive,whittington2017approximation,millidge2022predictive,salvatori2024stable} all retain some form of structured backward signal---contrasting with FF's complete elimination of backward gradient flow. Forward-only alternatives either modulate the input with the output error \citep{dellaferrera2022pepita} or estimate gradients from activity perturbations via forward-mode differentiation with local losses \citep{ren2023forward}. Auxiliary-classifier heads with local losses have a long history \citep{szegedy2015going,lee2015dsn,belilovsky2019greedy,belilovsky2020decoupled,nokland2019training}; our multi-layer classifier builds on this lineage.

\textbf{Information theory, normalization, and temperature scaling.}
Information-bottleneck perspectives on deep learning \citep{tishby1999information,tishby2015deep,shwartzziv2017opening} have been critiqued \citep{saxe2018information}, and high-dimensional MI estimation is itself fragile \citep{belghazi2018mine,poole2019variational,mcallester2020limitations}; we sidestep these issues by using histogram and $k$-nearest-neighbour estimators \citep{kraskov2004ksg,ross2014mi} for scalar MI and linear-probe lower bounds for vectors. Normalization's effect on optimization has been extensively studied \citep{ioffe2015batch,ba2016layer,santurkar2018does,yang2019mean}; our contribution is three-path decoupling rather than a new normalization. Temperature scaling appears in knowledge distillation \citep{hinton2015distilling}, calibration \citep{guo2017calibration}, and curriculum-based distillation \citep{li2023curriculum}, all operating on softmax outputs; TempBalance \citep{zhou2023tempbalance} instead treats layer-wise learning rates as temperatures. DTG differs from both by scaling the layer-local learning signal. Extended discussion in Appendix~\ref{app:related}.

\section{Diagnostic and Synthetic Validation}\label{sec:analysis}

\subsection{Per-Layer Signal Diagnostic}\label{sec:diag}

A short empirical diagnostic on a trained DTG-FF VGG8 (CIFAR-10, $91.33\%$) motivates the three DTG-FF components and grounds the BP-shadow lens of Sec.~\ref{sec:bp_shadow}. We measure scalar goodness $I(g_l^{\mathrm{scalar}};Y){\approx}0.24$ bits per layer (100-bin histogram, mean across layers, range $0.16$--$0.31$; $0.22$--$0.38$ bits across $50$--$500$ bins and $0.20$ bits with a $k$-nearest-neighbour estimator \citep{kraskov2004ksg,ross2014mi}), spatial goodness vectors at $1.1$--$2.5$ bits per layer via linear probe with Fano's inequality, and GAP features at $2.52$ bits at layer~$6$. Estimator details, bin-sensitivity, and the per-layer figure are in App.~\ref{app:mi_methodology}. FF's inter-layer gradient path is also \emph{zero by construction} via \texttt{detach}---a gradient-flow property, not an information-theoretic bound. These observations motivate three design pathways: \emph{signal quality} (scalar~$\to$~spatial goodness, providing higher probe-accessible task signal per layer), \emph{signal utilization} (dynamic temperature; ablation cost $-0.72$ to $-1.34$ pp at $T\!=\!1$), and \emph{cross-layer coordination} (multi-layer fusion; pairwise per-layer prediction disagreement $25.1\%$ on CIFAR-10 test, indicating non-redundant per-layer hypotheses to aggregate). We test these design choices in a controlled synthetic setup before committing to real-data scaling.

\subsection{Synthetic Validation with Architecture-Matched BP Controls}\label{sec:synthetic}

\textbf{Setup.}
A 3-layer ReLU teacher ($d_{\mathrm{in}}=50$, $d_{\mathrm{hidden}}=128$) labels 20{,}000 train / 5{,}000 test samples. All students have 4 hidden layers ($d_{\mathrm{hidden}}=128$), 8{,}000 Adam steps, batch $256$, $\mathrm{lr}=10^{-3}$, $5$ seeds, $K\in\{5,10,15,20,30,50\}$. We compare DTG-FF against single BP, BP-DeepSup (backbone- and depth-matched, differs from DTG-FF only in detach), and BP-Ensemble ($4\!\times$ params, softmax-averaged). Teacher variance dominates across seeds (std $4$--$17\%$), so we report paired differences. Full table, baseline definitions, and param counts are in App.~\ref{app:synth}.

\textbf{Results.} Paired DTG-FF $-$ BP-DeepSup grows from $-0.23$ pp at $K\!=\!5$ to $+2.00$ pp at $K\!=\!50$; a pre-specified low-$K$ ($\{5,10,15,20\}$) vs.\ high-$K$ ($\{30,50\}$) contrast yields $+1.37$ pp in favor of high $K$ (bootstrap $n\!=\!10{,}000$, $95\%$ CI $[+0.59, +2.15]$). DTG-FF reliably exceeds single BP at $K\!\geq\!15$ (5/5 seeds, $+1.3$ to $+3.4$ pp) but loses to the $4\!\times$-parameter BP-Ensemble at every $K$, as expected for a pure capacity comparison. The narrow finding: \emph{the FF--BP gap in this synthetic regime is not solely explained by the absence of end-to-end gradients}. The synthetic advantage does not transfer to real images: on CIFAR-10/100, architecture-matched BP-DeepSup outperforms DTG-FF by $2.40$/$5.93$ pp (Sec.~\ref{sec:sota}).

\section{Method: DTG-FF}\label{sec:method}

\subsection{Temperature-Scaled Local Goodness (DTG)}\label{sec:dtgmech}

Let layer $l$ compute activation $\mathbf{z}_l(x) = \mathrm{ReLU}(W_l \hat{\mathbf{x}}_{l-1})$ where $\hat{\mathbf{x}}_{l-1}$ is the (detached) input from layer $l\!-\!1$. We define a nonnegative goodness representation
\begin{equation}
	\mathbf{u}_l(x) = \phi_l(\mathbf{z}_l(x)) \in \mathbb{R}^{m_l}_{\geq 0},
\end{equation}
where $\phi_l$ depends on architecture: $\phi_l^{\mathrm{MLP}}(\mathbf{z}) = \tfrac{1}{d_l}\|\mathbf{z}\|_2^2 \in \mathbb{R}_{\geq 0}$ (scalar; $m_l\!=\!1$); $\phi_l^{\mathrm{CNN}}(\mathbf{z}) = \mathrm{flatten}(\mathrm{AdaptiveAvgPool2d}(\mathbf{z}^{2},(h_l,w_l))) \in \mathbb{R}^{m_l}_{\geq 0}$ (vector; here and in subsequent goodness expressions $\mathbf{z}^{2}$ denotes the element-wise square). DTG-FF introduces a layer-local learnable temperature $T_l > 0$ and uses the temperature-scaled goodness
\begin{equation}
	\tilde{\mathbf{u}}_l(x) = \mathbf{u}_l(x)/T_l, \qquad T_l = T_{\min} + (T_{\max}-T_{\min})\,\sigma(\alpha_l),
\end{equation}
where $\alpha_l\!\in\!\mathbb{R}$ is a per-layer learnable scalar (one parameter per layer) and $\sigma$ is the logistic sigmoid. Each layer therefore learns its own temperature jointly with its weights, via gradient descent on the layer-local loss.

\textbf{Locality.}
$T_l$ is a learnable parameter of layer $l$ and does not depend on other layers or batch statistics. DTG-FF's FF (goodness) loss therefore satisfies strict inter-layer gradient locality (enforced by \texttt{detach}): updates to $W_l$ require no gradient propagated through layers $j\!\neq\!l$. The MLP classifier head is a separate hybrid that backpropagates through shared layer weights (App.~\ref{app:locality}); the CNN classifier is fully detached from the conv backbone.

\begin{figure}[t]
	\centering
	\includegraphics[width=\textwidth]{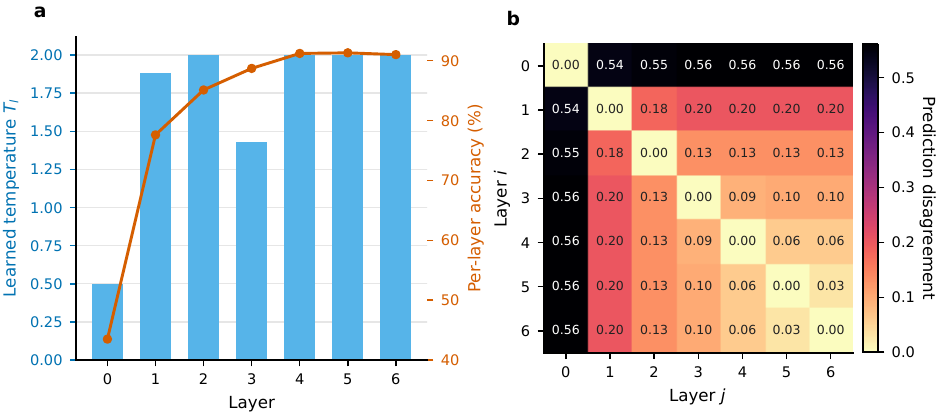}
	\caption{DTG mechanism diagnostics on the trained DTG-FF VGG8 \emph{concat-classifier} checkpoint (CIFAR-10, 91.33\%). \textbf{Left:} learned temperature $T_l$ per layer (bars, left axis) and per-layer random-projection classifier accuracy (line, right axis). Layer~0 learns $T\!=\!T_{\min}\!=\!0.5$; layers~1 and~3 take intermediate values ($1.9$ and $1.4$); layers~2 and~4--6 saturate at $T_{\max}\!=\!2.0$. \textbf{Right:} pairwise per-layer prediction disagreement on CIFAR-10 test. Layer~0 disagrees with deeper layers on 54--56\% of samples, whereas late layers are nearly redundant (layers~4--6 disagree by only 3--6\%); the off-diagonal mean disagreement is 25.1\%.}
	\label{fig:mechanism}
\end{figure}

\textbf{Learned temperatures.}
Figure~\ref{fig:mechanism} (left panel) shows that the learned $T_l$ in a trained DTG-FF VGG8 spans the full allowed range: layer~0 learns $T_l\!=\!T_{\min}\!=\!0.5$, layers~1 and~3 take intermediate values ($1.9$ and $1.4$), and layers~2 and~4--6 saturate at $T_{\max}\!=\!2.0$. The temperature range is used meaningfully rather than collapsing to a single value. At test time, the goodness loss is not computed; temperature still scales logits for the per-layer random-projection readout at inference (App.~\ref{app:ema}).

\textbf{Local loss.}
DTG-FF instantiates a shared temperature-scaled form $\mathcal{L}_l = \ell_l(\psi_l(\tilde{\mathbf{u}}_l), y)$ with two architecture-dependent readouts:

\emph{MLP: scalar margin loss.}
For positive/negative supervision $s\!\in\!\{+1,-1\}$ \citep{hinton2022forward},
\begin{equation}
	\psi_l^{\mathrm{MLP}}(\tilde u_l) = \tilde u_l - \theta_l,\quad \ell_l(\psi_l, s) = \log(1 + \exp(-s\,\psi_l)),
\end{equation}
where $\theta_l = \mathrm{softplus}(\theta_{0,l})$ is a learnable scalar margin per layer, initialized so that $\theta_l$ matches the average goodness measured on the first batch (App.~\ref{app:warmstart}).

\emph{CNN: $K$-way local cross-entropy.}
\begin{equation}
	\psi_l^{\mathrm{CNN}}(\tilde{\mathbf{u}}_l) = \tilde{\mathbf{u}}_l^\top R_l + \mathbf{b}_l,\quad \ell_l(\boldsymbol\psi, y) = \mathrm{CE}(\boldsymbol\psi, y),
\end{equation}
with fixed (non-learned) random projection $R_l\!\in\!\mathbb{R}^{m_l\times K}$, $R_{l,ij} \sim \mathcal{N}(0, 1/K)$, and $\mathbf{b}_l\!\sim\!\mathcal{N}(\mathbf{0}, (1/K)I)$ also fixed. No margin term is needed.

The choice of loss reflects convention: \citet{hinton2022forward} established scalar goodness + margin for MLPs; the FF-CNN literature \citep{papachristodoulou2024cwcomp,gong2026asge} uses vector goodness + cross-entropy. In the scalar-goodness case, a $K$-way linear readout $a_k = r_{lk} g_l/T_l + b_{lk}$ is rank-1 (all classes depend on the same scalar), and the margin objective circumvents this degeneracy; we do not claim this forces the design choice, but it explains why the two instantiations use different loss families.

\textbf{Gradient scale.}
For the CNN loss, $\nabla_{\mathbf{u}}\mathcal{L} = T^{-1} R_l(\mathbf{p}-\mathbf{e}_y)$, so $\|\nabla_{\mathbf{u}}\mathcal{L}\|_2 \!\leq\! (\sqrt{2}/T)\|R_l\|_{\mathrm{op}}$. With $T\!\in\![0.5,2.0]$ for CNN, DTG modulates gradient magnitude by at most $T_{\max}/T_{\min}\!=\!4$; for MLP with $T\!\in\![0.1,2.0]$ the range is $20$. This modulation does not change task-relevant mutual information $I(\mathbf{u};Y)$---temperature adapts the optimization signal, not the representation content. Appendix~\ref{app:rscale} discusses the random-projection scale confound and reports an ablation with column-normalized $R_l$.

\subsection{DTG-CNN: Decoupled Normalization and Multi-Layer Fusion}\label{sec:dtgcnn}

\textbf{Architecture.}
We use a VGG-style \citep{simonyan2015vgg} VGG8 backbone (7 convolutional layers: $3\!\to\!128\!\to\!256\!\to\!256\!\to\!512\!\to\!512\!\to\!512\!\to\!512$, with $2\!\times\!2$ average pooling after layers~1, 3, 4, 5). Each convolutional layer has three functional paths:

\emph{FF (goodness) path.} No normalization. The raw post-ReLU activation $\mathbf{z}_l$ feeds directly into spatial goodness $\mathbf{u}_l$ and the local loss $\mathcal{L}_l$. Batch normalization here collapses class-conditional variance in squared activation norms, reducing the signal the goodness objective relies on. Relatedly, \citet{sun2025deeperforward} use layer normalization with mean goodness and argue that CwComp's classification directly on batch-normalized outputs, without decoupling goodness, leaks goodness to the next layer.

\emph{Inter-layer propagation path.} Channel-wise LayerNorm applied at each spatial location (i.e., for $\mathbf{x}\!\in\!\mathbb{R}^{B\times C\times H\times W}$, normalize along $C$ per $(b,h,w)$), followed by dropout and \texttt{detach}. The LayerNorm module includes affine parameters $(\gamma, \beta)$ but its application sits inside a \texttt{torch.no\_grad} region in our pipelined per-layer training step (the FF goodness loss is computed on the pre-norm activation $\mathbf{z}_l$, and the propagated input $\mathbf{z}_{l+1}^{\text{in}}$ is derived only after the autograd graph has been freed); these affine parameters therefore receive no gradient in either path and remain at their initialization $(\gamma\!=\!1, \beta\!=\!0)$, so the path is operationally equivalent to non-affine LayerNorm. We retain the module form for code-level interchangeability with affine variants. This path provides the next convolutional filter with inputs of controlled scale, without affecting the FF training signal.

\emph{Classifier path.} Global Average Pooling per layer produces $\mathbf{f}_l \in \mathbb{R}^{C_l}$. Features from layers $1$ through $L\!-\!1$ (i.e., excluding the first conv layer; using $0$-based indexing this is layers $1, 2, \dots, L\!-\!1$) are concatenated, normalized by BatchNorm1d, passed through dropout, and fed to a learnable linear classifier trained with cross-entropy (label smoothing 0.1 \citep{szegedy2016rethinking}). For VGG8 ($L\!=\!7$) the concat dimension is $256+256+512+512+512+512 = 2560$. An alternative logit-sum aggregation (per-layer random-projection logits summed across \emph{all} $L$ layers, no trainable classifier head) is also evaluated and yields the best CIFAR-10 accuracy; Sec.~\ref{sec:ablation} compares both.

\textbf{Locality scope.}
Convolutional feature learning is fully inter-layer local: per-layer AdamW optimizers, per-layer cosine schedulers, \texttt{detach} at every layer boundary. The concatenated classifier head uses global backpropagation through BN and Linear, but not into the conv layers (features are extracted under \texttt{no\_grad}). The logit-sum variant requires no learnable classifier parameters, keeping the full pipeline layer-local. We follow the auxiliary-classifier tradition \citep{belilovsky2019greedy} in treating a global classifier head as a practical, well-understood concession.

\textbf{Training details.}
Per-layer AdamW \citep{loshchilov2019adamw} (lr $2\!\times\!10^{-4}$, weight decay $10^{-3}$) with cosine annealing \citep{loshchilov2017sgdr} to $10^{-5}$. 400 epochs for 32$\times$32 datasets; 200 for Tiny ImageNet and ImageNet-100. Augmentation: random crop (padding 4; random-resized crop for ImageNet-100), horizontal flip, color jitter, random grayscale ($p\!=\!0.2$), and Cutout \citep{devries2017cutout} ($16\!\times\!16$ for 32$\times$32, $56\!\times\!56$ for 224$\times$224). Classifier dropout 0.2, inter-layer dropout 0.1. Full hyperparameters are in App.~\ref{app:hyper}.

\section{Experiments}\label{sec:experiments}

\subsection{Setup}\label{sec:setup}

We evaluate DTG-FF on nine real-world classification datasets: CIFAR-10/100 \citep{krizhevsky2009learning}, Fashion-MNIST \citep{xiao2017fashion}, STL-10 \citep{coates2011analysis}, Tiny ImageNet \citep{le2015tiny}, ImageNet-100 (a 100-class subset of ImageNet \citep{deng2009imagenet}), and three MedMNIST datasets \citep{yang2023medmnist} (PathMNIST \citep{kather2019predicting}, DermaMNIST \citep{tschandl2018ham10000}, BloodMNIST \citep{acevedo2020dataset}). VGG8 is used for $32\!\times\!32$ inputs and for Tiny ImageNet at native $64\!\times\!64$ resolution; VGG11 is used only for ImageNet-100 ($224\!\times\!224$). Full training details in App.~\ref{app:hyper}; secondary datasets (STL-10, Fashion-MNIST, MedMNIST) are summarized in App.~\ref{app:secondary}.

\textbf{Checkpoint selection.} Unless stated otherwise, every accuracy we report for DTG-FF and for our own BP controls is the best accuracy on the official test split across the periodic evaluations made during training (the same rule for all methods). This is an optimistic, test-selected estimate rather than a held-out one; the FF--BP comparisons share the bias, but absolute numbers should be read with this caveat. Results quoted from prior work follow those papers' own protocols.

\subsection{FF-Family State of the Art}\label{sec:sota}

\begin{table}[t]
	\centering
	\small
	\setlength{\tabcolsep}{3.5pt}
	\renewcommand{\arraystretch}{0.95}
	\caption{FF-family comparison on CIFAR-10, CIFAR-100, Tiny ImageNet (200 classes), and ImageNet-100 ($224\!\times\!224$). Entries as reported; `---' = not reported. Bold marks the best result among methods public at our submission (May 2026).}\label{tab:sota}
	\begin{tabular}{@{}lccccc@{}}
		\toprule
		Method                                   & Arch  & CIFAR-10         & CIFAR-100        & Tiny-IN                      & IN-100           \\
		\midrule
		Hinton FF \citep{hinton2022forward}      & local. conn. & $\sim\!59$\%     & ---              & ---                          & ---              \\
		CwComp \citep{papachristodoulou2024cwcomp}            & CNN   & 78.11\%          & 51.23\%          & ---                          & ---              \\
		SoftHebb$^{\ast}$ \citep{journe2023softhebb} & CNN   & 80.31\%          & ---              & ---                          & ---              \\
		SCFF$^{\ast}$ \citep{chen2025scff}          & CNN   & 80.75\%          & ---              & 35.67\%                      & ---              \\
		DeeperForward \citep{sun2025deeperforward}  & ResNet-17 & 88.72\%$^{\S}$ & 53.09\%          & ---                          & ---              \\
		ASGE-VGG8 \citep{gong2026asge}           & VGG8  & 90.50\%          & 65.34\%          & ---                          & ---              \\
		ASGE-VGG11 \citep{gong2026asge}          & VGG11 & 90.62\%          & 65.42\%          & ---                          & ---              \\
		\midrule
		\multicolumn{6}{@{}l}{\emph{Concurrent work, public after our submission:}}                                                           \\
		HCL-FF \citep{yao2026hclff}              & ResNet-17 & 91.68\%      & 70.09\%          & 48.46\%                      & ---              \\
		BiCovG+FAL \citep{jiang2026bicovg}       & VGG8  & ---              & 70.80\%          & ---                          & ---              \\
		BiCovG+FAL \citep{jiang2026bicovg}       & VGG-16 & ---             & ---              & 50.30\%                      & 73.01\%          \\
		Forward-only CNN \citep{ghader2026forwardonly} & ResNet-17 & 89.37\% & 63.52\%         & 49.87\%                      & ---              \\
		\midrule
		\textbf{DTG-FF (concat)}                 & VGG8  & \textbf{91.33\%} & \textbf{67.28\%} & \textbf{48.17\%}$^{\dagger}$ & ---              \\
		\textbf{DTG-FF (logit-sum)}              & VGG8  & \textbf{91.79\%} & ---              & ---                          & ---              \\
		\textbf{DTG-FF}                          & VGG11 & ---              & ---              & ---                          & \textbf{49.40\%} \\
		\midrule
		\multicolumn{6}{@{}l}{\emph{Backpropagation references (ours, same VGG8 backbone, same training recipe):}}                               \\
		BP-VGG8-DeepSup (arch-matched to DTG-FF) & VGG8  & 93.73\%          & 73.21\%          & ---                          & ---              \\
		BP-VGG8 (concat classifier, single head) & VGG8  & 94.80\%          & ---              & ---                          & ---              \\
		\bottomrule
	\end{tabular}\\
	\footnotesize{$^{\ast}$Unsupervised training followed by a linear readout; SoftHebb is Hebbian rather than FF. $^{\S}$1000 epochs with stronger augmentation; $86.22\%$ in the main 150-epoch protocol. $^{\dagger}$Tiny ImageNet at native $64\!\times\!64$ resolution. ASGE also reports $51.58\%$ top-1 on ImageNet-1k with a $4\times$-wide ResNet-18 \citep{gong2026asge}; BiCovG feeds $224\!\times\!224$ ImageNet-100 images downsampled to $112\!\times\!112$ to its first layer. Notably, on tractable texture-based medical classification DTG-FF reaches \textbf{96.61\%} on BloodMNIST, within the typical BP range; full secondary-benchmark results (Fashion-MNIST, STL-10, PathMNIST, DermaMNIST) in App.~\ref{app:secondary}.}
\end{table}

Table~\ref{tab:sota} summarizes FF-family results. DTG-FF (concat) exceeded the prior best FF-family CIFAR-10 result at the time of our submission by $+0.71\%$ with a smaller backbone (VGG8 vs.\ ASGE's VGG11), on CIFAR-100 by $+1.86\%$, and achieves 48.17\% on Tiny ImageNet---well above the next-highest FF-family result public at that time ($35.67\%$, SCFF). Concurrent work released after our submission reports higher CIFAR-100 and Tiny ImageNet results and a comparable CIFAR-10 result \citep{yao2026hclff,jiang2026bicovg,ghader2026forwardonly}; we list it separately in Table~\ref{tab:sota}. An alternative logit-sum aggregation variant (no trainable classifier head; per-layer random-projection logits summed at inference) reaches 91.79\% on CIFAR-10; we report this as a CIFAR-10-specific ablation rather than a cross-dataset headline, since we evaluate only the concat variant on the remaining datasets.

\textbf{Architecture-matched BP comparison.}
We report two BP reference points on the same VGG8 backbone and identical training recipe (AdamW, lr $2\!\times\!10^{-4}$, cosine to $10^{-5}$, 400 epochs, same augmentation): (i) \textbf{BP-VGG8-DeepSup}, end-to-end BP with per-layer auxiliary heads and logit-sum inference---matched to DTG-FF in backbone, depth, and per-layer head count, but differing in detach boundary, local objective, and head parameterization (see Sec.~\ref{sec:limitations}) (93.73\% / 73.21\% on CIFAR-10/100); (ii) \textbf{BP-VGG8}, a single-head classifier (94.80\% on CIFAR-10). The FF--BP gap on the concat operating point is $\mathbf{2.40}$ pp on CIFAR-10 and $\mathbf{5.93}$ pp on CIFAR-100; under strict aggregation matching (DTG-FF logit-sum $91.79\%$ vs.\ BP-DeepSup logit-sum), the CIFAR-10 gap is $1.94$ pp (CIFAR-100 logit-sum not reported). The gap widens with task complexity in either framing.

\textbf{Multi-seed stability.}
Headline numbers use seed 42; reproducing on seed 43 gives DTG-FF $91.18$/$66.95\%$ and BP-DeepSup $93.63$/$72.80\%$, with paired per-seed gaps $\{2.40, 2.45\}$ pp on CIFAR-10 and $\{5.93, 5.85\}$ pp on CIFAR-100. Both methods reproduce within $0.4$ pp std and the FF--BP gap is stable to $0.05$/$0.08$ pp.

\begin{figure}[t]
	\centering
	\includegraphics[width=0.78\textwidth]{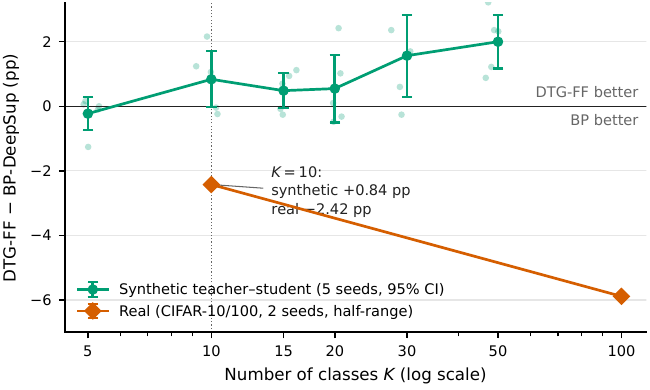}
	\caption{\textbf{Synthetic vs.\ real $K$-axis conflict.} Paired DTG-FF $-$ BP-DeepSup accuracy difference on a shared $K$-axis. \textbf{Synthetic} (green, 5 teacher--student seeds): DTG-FF advantage \emph{grows} with $K$, reaching $+2$ pp at $K\!=\!50$. \textbf{Real} (orange, CIFAR-10/100, 2 seeds, half-range error bars): the gap is reversed and \emph{widens} from $-2.42$ pp at $K\!=\!10$ to $-5.89$ pp at $K\!=\!100$. At the matched $K\!=\!10$ point the synthetic regime predicts a $+0.84$ pp DTG-FF advantage; the real-data outcome is $-2.42$ pp---a $3.3$ pp sign-reversed disagreement on identical class count. This visualization motivates the methodology critique of Sec.~\ref{sec:granularity}: synthetic $K$-sweeps confound output dimensionality with fine-grained discrimination difficulty and overstate FF transferability.}
	\label{fig:synthetic_real_k}
\end{figure}

\textbf{ImageNet-100 ceiling.}
At $224\!\times\!224$ on ImageNet-100, DTG-FF on VGG11 reaches $49.4\%$ in $200$ epochs ($12.6$ hours on a single RTX~4090)---ASGE \citep{gong2026asge} previously reported FF training on ImageNet-1k, and concurrent BiCovG reports $73.01\%$ on ImageNet-100 with VGG-16 \citep{jiang2026bicovg}---trailing the $75.6\%$ that CMC reports for a ResNet-50 even without labels (self-supervised pre-training with linear evaluation) \citep{tian2020cmc} by a wide margin. That reference uses ResNet-50, so the precise gap conflates algorithmic and architectural capacity; the conservative reading is that DTG-FF at $224\!\times\!224$ remains substantially below standard BP's operating range, with the per-architecture penalty left to a future BP-VGG11 control.

\textbf{MLP and depth-scaling supplementaries.}
DTG also improves FF-MLP on CIFAR-10 to $63.72\%$, above the $\sim\!59\%$ reported for the original FF on locally connected networks \citep{hinton2022forward}, although our MLP uses a hybrid BP-trained classifier head (App.~\ref{app:mlp}, App.~\ref{app:locality}). A deeper VGG11 backbone on CIFAR-10/100 underperforms VGG8 by $6.97$/$12.49$ pp respectively (App.~\ref{app:depth}); VGG11 also differs in early-channel width (64 vs.\ 128) and downsampling, so this bundles depth and width effects. We retain VGG11 only for ImageNet-100, where higher per-sample signal partially compensates.

\subsection{Within-Dataset $K$ Probe: CIFAR-100 Coarse vs Fine}\label{sec:granularity}\label{sec:imagenet100}

The cross-dataset CIFAR-10/CIFAR-100 comparison confounds class count with image distribution. CIFAR-100 supplies both fine ($K\!=\!100$) and coarse ($K\!=\!20$, semantic superclass) labels over the \emph{same} images \citep{krizhevsky2009learning}. At matched recipe (VGG8, $50$ epochs, seed $114514$; App.~\ref{app:granularity}) the FF--BP gap shrinks from $11.35$ pp (fine) to $7.99$ pp (coarse)---a $3.36$ pp differential. Both runs are non-converged at $50$ epochs and the fine gap drifts $5.42$ pp by $400$ epochs (exceeding the $3.36$ pp differential), so the probe is direction-suggestive rather than asymptotic; matched-convergence is left to future work.

\subsection{Ablation on CIFAR-10}\label{sec:ablation}

\begin{table}[h]
	\centering
	\small
	\setlength{\tabcolsep}{6pt}
	\renewcommand{\arraystretch}{0.95}
	\caption{CIFAR-10 ablation, headline rows (VGG8, 400 epochs, single seed). Cutout, RMSNorm, and regularization variants in App.~\ref{app:ablation}.}\label{tab:ablation}
	\begin{tabular}{@{}lcr@{}}
		\toprule
		Configuration                    & Accuracy         & $\Delta$ \\
		\midrule
		\textbf{DTG-FF (logit-sum)}      & \textbf{91.79\%} & ---      \\
		\ \ $-$ DTG (fixed $T\!=\!1.0$)  & 90.45\%          & $-1.34$  \\
		\midrule
		DTG-FF (concat)                  & 91.33\%          & ---      \\
		\ \ $-$ DTG (fixed $T\!=\!1.0$)  & 90.61\%          & $-0.72$  \\
		\ \ $-$ inter-layer LayerNorm    & 90.53\%          & $-0.80$  \\
		\bottomrule
	\end{tabular}
\end{table}

Removing DTG (fixed $T\!=\!1$) is the dominant single-component effect, consistent across aggregation schemes ($-0.72$ pp concat, $-1.34$ pp logit-sum). Removing inter-layer LayerNorm contributes $-0.80$ pp; replacing it with RMSNorm \citep{zhang2019rmsnorm} costs $-0.18$ pp (at 200 epochs). Cutout and explicit regularization (label smoothing $+$ classifier dropout) have marginal effect on final accuracy under cosine annealing, though they substantially affect training-loss trajectories (App.~\ref{app:ablation}).

\section{Discussion and Limitations}\label{sec:discussion}

\subsection{A BP-Shadow Lens: FF Improvements as Partial Substitutes}\label{sec:bp_shadow}

We propose a unifying lens for our four scaling observations: several established FF-family improvements can be interpreted as partial substitutes for the supervised cross-layer signal that BP provides natively, and the residual architecture-matched gap reflects coordination and supervision that no current substitute fully recovers. The architecture-matched audit (Sec.~\ref{sec:sota}, App.~\ref{app:depth}) provides empirical grounding for this reading.

\begin{center}
	\small
	\setlength{\tabcolsep}{4pt}
	\renewcommand{\arraystretch}{1.0}
	\begin{tabular}{@{}p{0.30\linewidth}p{0.32\linewidth}p{0.32\linewidth}@{}}
		\toprule
		FF-family component                                                              & Substitute role                                          & BP equivalent                                          \\
		\midrule
		Label overlay \citep{hinton2022forward}                                          & Input-side label injection                               & Backward label gradient at every layer                 \\
		BP-trained classifier head (hybrid FF)                                           & Adds supervised cross-layer gradient in hybrid variants   & End-to-end supervised gradient                        \\
		Spatial goodness vector \citep{gong2026asge}                      & Wider local learning signal per layer                    & Richer per-layer learning signal                       \\
		Multi-layer fusion (this work, ASGE \citep{gong2026asge})                        & Post-hoc cross-layer aggregation                         & In-training cross-layer gradient flow                  \\
		\bottomrule
	\end{tabular}\\
	\footnotesize{Mappings are interpretive: each row pairs an FF-family mechanism with a BP information channel it appears to substitute for. The mappings are not causally validated by these measurements alone (Sec.~\ref{sec:limitations}).}
\end{center}

Under this lens, BP-DeepSup's matched advantage of $2.40$/$5.93$ pp, the larger real-data gap at higher $K$, the depth degradation, and the synthetic--real $K$ reversal are coherent consequences of incomplete substitutes for supervised cross-layer coordination. Two further FF mechanisms---BatchNorm in goodness-adjacent paths and the dynamic temperature in this work---are not obvious substitutes for any single BP channel; we treat them as adjacent rather than load-bearing for this lens. Causally isolating which substitute carries the dominant load requires controls that swap one channel at a time (Sec.~\ref{sec:limitations}).

\subsection{Auditing the Systems Advantage of Layer-Local Training}\label{sec:systems}

A natural defense of FF research is its structural systems property: pipelined layer-local training admits an $\mathcal{O}(1)$-in-depth activation-memory bound, whereas BP saves activations across all $L$ layers. We measure this directly to test whether it translates into practical dominance. Implementation matters: a naive forward-once-with-grad / forward-once-no-grad split duplicates conv computation and underperforms BP-VGG8 by $5\%$ memory and $30\%$ throughput; a pipelined schedule (single conv pass, autograd graph released by local backward before deriving the next-layer input) recovers the structural property. Within VRAM (VGG8/$32\!\times\!32$ batch $256$), pipelined DTG-FF reaches $1.535$ GB / $1726$ imgs/s versus BP-VGG8 at $1.589$ GB / $1586$ imgs/s ($-3.4\%$ memory, $+9\%$ throughput). The harder test is the memory-cliff regime, VGG11/$224\!\times\!224$ on a commodity 8\,GB GPU, where we compare four methods at effective batch $128$ (full per-batch sweep and protocol in App.~\ref{app:memory}):

\begin{center}
	\small
	\setlength{\tabcolsep}{6pt}
	\begin{tabular}{@{}lccclc@{}}
		\toprule
		Method                              & Phys.\ batch & Peak GB    & imgs/s     & Extra cost                   & Source \\
		\midrule
		Vanilla BP-VGG11 (b=128)            & $128$        & $8.18^{*}$ & $14^{*}$           & host-spill                   & meas.  \\
		\textbf{BP+grad-accum} ($64\!\times\!2$) & $64$    & $4.18$     & $157^{\dagger}$    & $2\!\times$ fwd/bwd passes    & meas.  \\
		BP+grad-accum ($32\!\times\!4$)     & $32$         & $2.19$     & $157^{\dagger}$    & $4\!\times$ passes            & meas.  \\
		BP+activation-ckpt (b=128)          & $128$        & $6.35$     & $92$               & per-block recompute           & meas.  \\
		DTG-FF (b=128)                      & $128$        & $7.90$     & $138$              & none                          & meas.  \\
		\bottomrule
	\end{tabular}\\
	\footnotesize{${}^{*}$Vanilla BP at b=128 spills to host memory; remaining rows fit in VRAM. RTX 4060 Laptop, $8$ GB. ${}^{\dagger}$Identical throughput at b=64 and b=32 reflects compute-bound saturation on this GPU: per-step latency scales linearly with micro-batch size while effective imgs/s stays constant.}
\end{center}

\textbf{Finding.} On commodity 8\,GB hardware, the structural $\mathcal{O}(L)\!\to\!\mathcal{O}(1)$ activation-memory property of pipelined FF is realized but does not translate into measured systems dominance over memory-optimized BP. BP+gradient-accumulation at micro-batch $64$ matches DTG-FF's effective batch with $-47\%$ peak memory and $+14\%$ throughput; BP+activation-checkpointing \citep{chen2016checkpoint} recovers in-VRAM operation at a $1.5\!\times$ slowdown relative to DTG-FF; DTG-FF dominates only vanilla BP at its spill point---a regime practitioners avoid via standard tooling. The systems-feasibility argument for FF research at this scale is, on this hardware, not supported under fair baselines. Whether the structural property becomes load-bearing in deeper architectures or multi-device pipelines is an open question (App.~\ref{app:memory}).

\subsection{Limitations and Other Notes}\label{sec:limitations}

\textbf{Causal underidentification.}
Our scaling diagnosis compares DTG-FF against BP-DeepSup, which differs along three per-layer dimensions (detach boundary, goodness vs.\ CE objective, fixed random projection vs.\ learned head). The architecture-matched gap is real, but cleanly isolating which dimension dominates requires a sequence of single-dimension ablations (e.g., a Local-CE-Detach control to isolate detach, plus learned-head DTG and detached BP-DeepSup variants for the remaining two dimensions) left to future work. The BP-shadow lens (Sec.~\ref{sec:bp_shadow}) is therefore interpretive synthesis, not causal validation.

\textbf{Ensemble scope.}
A BP ensemble with $4\!\times$ parameters exceeds DTG-FF on synthetic tasks at every $K$. DTG-FF is a local-learning mechanism, not a parameter-efficiency argument; claims about FF-family advantages are regime-qualified.

\textbf{Relationship to DeeperForward.}
\citet{sun2025deeperforward} replace FF's length normalization with layer normalization and mean goodness, and argue that CwComp's classification directly on batch-normalized outputs, without decoupling goodness, leaks goodness to the next layer. We decouple normalization across functional paths (Sec.~\ref{sec:dtgcnn}): no normalization on the goodness path, LayerNorm on the propagation path and BN only in the detached classifier.

\textbf{Biological plausibility.}
DTG-FF uses BatchNorm in the classifier, dropout, and AdamW; learnable temperatures are engineering parameters. We describe it as \emph{biologically motivated} (layer-local gradient flow) rather than strictly biologically plausible.

\section{Conclusion}\label{sec:conclusion}

We developed DTG-FF as an FF-family instrument that set state of the art across nine benchmarks at the time of our submission (CIFAR-10/100, Tiny ImageNet, an ImageNet-100 baseline at $224\!\times\!224$, and others), and used it as a stress test for strictly layer-local training. Even at FF-family SOTA, DTG-FF trails an architecture-matched BP-DeepSup baseline under identical recipe by $2.40$/$5.93$ pp on CIFAR-10/100, with the gap widening in class count; on ImageNet-100 at $224\!\times\!224$ it reaches $49.4\%$ against $75.6\%$ even for self-supervised BP with linear evaluation~\citep{tian2020cmc}, a real-data ceiling that persists at higher resolution. A within-dataset CIFAR-100 coarse-vs-fine probe combined with the synthetic-vs-real $K$-axis reversal supports an interpretation in which synthetic $K$-sweeps confound output dimensionality with fine-grained discrimination, leading current FF benchmarks to overstate real-data transferability. A unifying lens reads several FF-family improvements as partial substitutes for the supervised cross-layer signal that BP provides natively; under fair systems baselines, even the structural $\mathcal{O}(1)$-in-depth activation property of pipelined FF does not translate into measured advantage over BP+gradient-accumulation on commodity hardware.

\bibliography{references}

@article{hinton2022forward,
  title={The Forward-Forward Algorithm: Some Preliminary Investigations},
  author={Hinton, Geoffrey},
  journal={arXiv preprint arXiv:2212.13345},
  year={2022}
}

@inproceedings{papachristodoulou2024cwcomp,
  title={Convolutional Channel-wise Competitive Learning for the Forward-Forward Algorithm},
  author={Papachristodoulou, Andreas and Kyrkou, Christos and Timotheou, Stelios and Theocharides, Theocharis},
  booktitle={Proceedings of the AAAI Conference on Artificial Intelligence},
  volume={38},
  year={2024},
  doi={10.1609/aaai.v38i13.29369}
}

@inproceedings{journe2023softhebb,
  title={Hebbian Deep Learning Without Feedback},
  author={Journ{\'e}, Adrien and Rodriguez, Hector Garcia and Guo, Qinghai and Moraitis, Timoleon},
  booktitle={International Conference on Learning Representations},
  year={2023}
}

@article{lee2023symba,
  title={{SymBa}: Symmetric Backpropagation-Free Contrastive Learning with Forward-Forward Algorithm for Optimizing Convergence},
  author={Lee, Heung-Chang and Song, Jeonggeun},
  journal={arXiv preprint arXiv:2303.08418},
  year={2023}
}

@article{chen2025scff,
  title={Self-Contrastive Forward-Forward Algorithm},
  author={Chen, Xing and Liu, Dongshu and Laydevant, J{\'e}r{\'e}mie and Grollier, Julie},
  journal={Nature Communications},
  volume={16},
  pages={5978},
  year={2025},
  doi={10.1038/s41467-025-61037-0}
}

@inproceedings{sun2025deeperforward,
  title={{DeeperForward}: Enhanced Forward-Forward Training for Deeper and Better Performance},
  author={Sun, Liang and Zhang, Yang and He, Weizhao and Wen, Jiajun and Shen, Linlin and Xie, Weicheng},
  booktitle={International Conference on Learning Representations},
  year={2025}
}

@article{gong2026asge,
  title={Adaptive Spatial Goodness Encoding: Advancing and Scaling Forward-Forward Learning Without Backpropagation},
  author={Gong, Qingchun and Staszewski, Robert Bogdan and Xu, Kai},
  journal={arXiv preprint arXiv:2509.12394},
  note={v2; accepted to ICASSP 2026},
  year={2026}
}

@article{ororbia2023pff,
  title={The Predictive Forward-Forward Algorithm},
  author={Ororbia, Alexander and Mali, Ankur},
  journal={arXiv preprint arXiv:2301.01452},
  year={2023}
}

@article{crick1989,
  title={The Recent Excitement about Neural Networks},
  author={Crick, Francis},
  journal={Nature},
  volume={337},
  pages={129--132},
  year={1989},
  doi={10.1038/337129a0}
}

@article{lillicrap2020backprop,
  title={Backpropagation and the Brain},
  author={Lillicrap, Timothy P. and Santoro, Adam and Marris, Luke and Akerman, Colin J. and Hinton, Geoffrey},
  journal={Nature Reviews Neuroscience},
  volume={21},
  number={6},
  pages={335--346},
  year={2020},
  doi={10.1038/s41583-020-0277-3}
}

@article{lillicrap2016feedback,
  title={Random Synaptic Feedback Weights Support Error Backpropagation for Deep Learning},
  author={Lillicrap, Timothy P. and Cownden, Daniel and Tweed, Douglas B. and Akerman, Colin J.},
  journal={Nature Communications},
  volume={7},
  pages={13276},
  year={2016},
  doi={10.1038/ncomms13276}
}

@inproceedings{nokland2016direct,
  title={Direct Feedback Alignment Provides Learning in Deep Neural Networks},
  author={N{\o}kland, Arild},
  booktitle={Advances in Neural Information Processing Systems},
  volume={29},
  year={2016}
}

@inproceedings{launay2020dfa,
  title={Direct Feedback Alignment Scales to Modern Deep Learning Tasks and Architectures},
  author={Launay, Julien and Poli, Iacopo and Boniface, Fran{\c{c}}ois and Krzakala, Florent},
  booktitle={Advances in Neural Information Processing Systems},
  volume={33},
  pages={9346--9360},
  year={2020}
}

@article{bengio2014target,
  title={How Auto-Encoders Could Provide Credit Assignment in Deep Networks via Target Propagation},
  author={Bengio, Yoshua},
  journal={arXiv preprint arXiv:1407.7906},
  year={2014}
}

@inproceedings{lee2015target,
  title={Difference Target Propagation},
  author={Lee, Dong-Hyun and Zhang, Saizheng and Fischer, Asja and Bengio, Yoshua},
  booktitle={Machine Learning and Knowledge Discovery in Databases (ECML PKDD 2015)},
  series={Lecture Notes in Computer Science},
  volume={9284},
  pages={498--515},
  publisher={Springer},
  year={2015},
  doi={10.1007/978-3-319-23528-8_31}
}

@inproceedings{meulemans2020theoretical,
  title={A Theoretical Framework for Target Propagation},
  author={Meulemans, Alexander and Carzaniga, Francesco S. and Suykens, Johan A. K. and Sacramento, Jo{\~a}o and Grewe, Benjamin F.},
  booktitle={Advances in Neural Information Processing Systems},
  volume={33},
  year={2020}
}

@article{scellier2017equilibrium,
  title={Equilibrium Propagation: Bridging the Gap Between Energy-Based Models and Backpropagation},
  author={Scellier, Benjamin and Bengio, Yoshua},
  journal={Frontiers in Computational Neuroscience},
  volume={11},
  pages={24},
  year={2017},
  doi={10.3389/fncom.2017.00024}
}

@inproceedings{laborieux2024improving,
  title={Improving Equilibrium Propagation without Weight Symmetry through {J}acobian Homeostasis},
  author={Laborieux, Axel and Zenke, Friedemann},
  booktitle={International Conference on Learning Representations},
  year={2024}
}

@inproceedings{millidge2023backpropagation,
  title={Backpropagation at the Infinitesimal Inference Limit of Energy-Based Models: Unifying Predictive Coding, Equilibrium Propagation, and Contrastive {H}ebbian Learning},
  author={Millidge, Beren and Song, Yuhang and Salvatori, Tommaso and Lukasiewicz, Thomas and Bogacz, Rafal},
  booktitle={International Conference on Learning Representations},
  year={2023},
  url={https://openreview.net/forum?id=nIMifqu2EO}
}

@article{rao1999predictive,
  title={Predictive Coding in the Visual Cortex: A Functional Interpretation of Some Extra-Classical Receptive-Field Effects},
  author={Rao, Rajesh P. N. and Ballard, Dana H.},
  journal={Nature Neuroscience},
  volume={2},
  number={1},
  pages={79--87},
  year={1999},
  doi={10.1038/4580}
}

@article{whittington2017approximation,
  title={An Approximation of the Error Backpropagation Algorithm in a Predictive Coding Network with Local {H}ebbian Synaptic Plasticity},
  author={Whittington, James C. R. and Bogacz, Rafal},
  journal={Neural Computation},
  volume={29},
  number={5},
  pages={1229--1262},
  year={2017},
  doi={10.1162/NECO_a_00949}
}

@article{millidge2022predictive,
  title={Predictive Coding Approximates Backprop Along Arbitrary Computation Graphs},
  author={Millidge, Beren and Tschantz, Alexander and Buckley, Christopher L.},
  journal={Neural Computation},
  volume={34},
  number={6},
  pages={1329--1368},
  year={2022},
  doi={10.1162/neco_a_01497}
}

@inproceedings{salvatori2024stable,
  title={A Stable, Fast, and Fully Automatic Learning Algorithm for Predictive Coding Networks},
  author={Salvatori, Tommaso and Song, Yuhang and Yordanov, Yordan and Millidge, Beren and Sha, Lei and Emde, Cornelius and Xu, Zhenghua and Bogacz, Rafal and Lukasiewicz, Thomas},
  booktitle={International Conference on Learning Representations},
  year={2024},
  url={https://openreview.net/forum?id=RyUvzda8GH}
}

@inproceedings{dellaferrera2022pepita,
  title={Error-driven Input Modulation: Solving the Credit Assignment Problem without a Backward Pass},
  author={Dellaferrera, Giorgia and Kreiman, Gabriel},
  booktitle={Proceedings of the 39th International Conference on Machine Learning},
  series={Proceedings of Machine Learning Research},
  volume={162},
  pages={4937--4955},
  publisher={PMLR},
  year={2022}
}

@inproceedings{ren2023forward,
  title={Scaling Forward Gradient With Local Losses},
  author={Ren, Mengye and Kornblith, Simon and Liao, Renjie and Hinton, Geoffrey},
  booktitle={International Conference on Learning Representations},
  year={2023},
  url={https://openreview.net/forum?id=JxpBP1JM15-}
}

@inproceedings{tishby1999information,
  title={The Information Bottleneck Method},
  author={Tishby, Naftali and Pereira, Fernando C. and Bialek, William},
  booktitle={Proceedings of the 37th Annual Allerton Conference on Communication, Control, and Computing},
  year={1999},
  note={arXiv:physics/0004057}
}

@inproceedings{tishby2015deep,
  title={Deep Learning and the Information Bottleneck Principle},
  author={Tishby, Naftali and Zaslavsky, Noga},
  booktitle={2015 IEEE Information Theory Workshop (ITW)},
  pages={1--5},
  year={2015},
  doi={10.1109/ITW.2015.7133169}
}

@article{shwartzziv2017opening,
  title={Opening the Black Box of Deep Neural Networks via Information},
  author={Shwartz-Ziv, Ravid and Tishby, Naftali},
  journal={arXiv preprint arXiv:1703.00810},
  year={2017}
}

@inproceedings{saxe2018information,
  title={On the Information Bottleneck Theory of Deep Learning},
  author={Saxe, Andrew M. and Bansal, Yamini and Dapello, Joel and Advani, Madhu and Kolchinsky, Artemy and Tracey, Brendan D. and Cox, David D.},
  booktitle={International Conference on Learning Representations},
  year={2018}
}

@inproceedings{belghazi2018mine,
  title={Mutual Information Neural Estimation},
  author={Belghazi, Mohamed Ishmael and Baratin, Aristide and Rajeswar, Sai and Ozair, Sherjil and Bengio, Yoshua and Courville, Aaron and Hjelm, R. Devon},
  booktitle={Proceedings of the 35th International Conference on Machine Learning},
  series={Proceedings of Machine Learning Research},
  volume={80},
  pages={531--540},
  year={2018}
}

@inproceedings{poole2019variational,
  title={On Variational Bounds of Mutual Information},
  author={Poole, Ben and Ozair, Sherjil and van den Oord, A{\"a}ron and Alemi, Alexander A. and Tucker, George},
  booktitle={Proceedings of the 36th International Conference on Machine Learning},
  series={Proceedings of Machine Learning Research},
  volume={97},
  pages={5171--5180},
  year={2019}
}

@inproceedings{mcallester2020limitations,
  title={Formal Limitations on the Measurement of Mutual Information},
  author={McAllester, David and Stratos, Karl},
  booktitle={Proceedings of the 23rd International Conference on Artificial Intelligence and Statistics},
  series={Proceedings of Machine Learning Research},
  volume={108},
  pages={875--884},
  year={2020}
}

@article{kraskov2004ksg,
  title={Estimating Mutual Information},
  author={Kraskov, Alexander and St{\"o}gbauer, Harald and Grassberger, Peter},
  journal={Physical Review E},
  volume={69},
  number={6},
  pages={066138},
  year={2004},
  doi={10.1103/PhysRevE.69.066138}
}

@inproceedings{szegedy2015going,
  title={Going Deeper with Convolutions},
  author={Szegedy, Christian and Liu, Wei and Jia, Yangqing and Sermanet, Pierre and Reed, Scott and Anguelov, Dragomir and Erhan, Dumitru and Vanhoucke, Vincent and Rabinovich, Andrew},
  booktitle={Proceedings of the IEEE Conference on Computer Vision and Pattern Recognition (CVPR)},
  pages={1--9},
  year={2015},
  doi={10.1109/CVPR.2015.7298594}
}

@inproceedings{lee2015dsn,
  title={Deeply-Supervised Nets},
  author={Lee, Chen-Yu and Xie, Saining and Gallagher, Patrick and Zhang, Zhengyou and Tu, Zhuowen},
  booktitle={Proceedings of the 18th International Conference on Artificial Intelligence and Statistics},
  series={Proceedings of Machine Learning Research},
  volume={38},
  pages={562--570},
  year={2015}
}

@inproceedings{belilovsky2019greedy,
  title={Greedy Layerwise Learning Can Scale to {ImageNet}},
  author={Belilovsky, Eugene and Eickenberg, Michael and Oyallon, Edouard},
  booktitle={Proceedings of the 36th International Conference on Machine Learning},
  series={Proceedings of Machine Learning Research},
  volume={97},
  pages={583--593},
  year={2019}
}

@inproceedings{belilovsky2020decoupled,
  title={Decoupled Greedy Learning of {CNNs}},
  author={Belilovsky, Eugene and Eickenberg, Michael and Oyallon, Edouard},
  booktitle={Proceedings of the 37th International Conference on Machine Learning},
  series={Proceedings of Machine Learning Research},
  volume={119},
  pages={736--745},
  year={2020}
}

@inproceedings{nokland2019training,
  title={Training Neural Networks with Local Error Signals},
  author={N{\o}kland, Arild and Eidnes, Lars Hiller},
  booktitle={Proceedings of the 36th International Conference on Machine Learning},
  series={Proceedings of Machine Learning Research},
  volume={97},
  pages={4839--4850},
  year={2019}
}

@inproceedings{ioffe2015batch,
  title={Batch Normalization: Accelerating Deep Network Training by Reducing Internal Covariate Shift},
  author={Ioffe, Sergey and Szegedy, Christian},
  booktitle={International Conference on Machine Learning},
  year={2015}
}

@inproceedings{santurkar2018does,
  title={How Does Batch Normalization Help Optimization?},
  author={Santurkar, Shibani and Tsipras, Dimitris and Ilyas, Andrew and Madry, Aleksander},
  booktitle={Advances in Neural Information Processing Systems},
  volume={31},
  year={2018}
}

@article{ba2016layer,
  title={Layer Normalization},
  author={Ba, Jimmy Lei and Kiros, Jamie Ryan and Hinton, Geoffrey E.},
  journal={arXiv preprint arXiv:1607.06450},
  year={2016}
}

@inproceedings{yang2019mean,
  title={A Mean Field Theory of Batch Normalization},
  author={Yang, Greg and Pennington, Jeffrey and Rao, Vinay and Sohl-Dickstein, Jascha and Schoenholz, Samuel S.},
  booktitle={International Conference on Learning Representations},
  year={2019}
}

@article{hinton2015distilling,
  title={Distilling the Knowledge in a Neural Network},
  author={Hinton, Geoffrey and Vinyals, Oriol and Dean, Jeff},
  journal={arXiv preprint arXiv:1503.02531},
  note={NIPS 2014 Deep Learning Workshop},
  year={2015}
}

@inproceedings{guo2017calibration,
  title={On Calibration of Modern Neural Networks},
  author={Guo, Chuan and Pleiss, Geoff and Sun, Yu and Weinberger, Kilian Q.},
  booktitle={International Conference on Machine Learning},
  year={2017}
}

@inproceedings{li2023curriculum,
  title={Curriculum Temperature for Knowledge Distillation},
  author={Li, Zheng and Li, Xiang and Yang, Lingfeng and Zhao, Borui and Song, Renjie and Luo, Lei and Li, Jun and Yang, Jian},
  booktitle={Proceedings of the AAAI Conference on Artificial Intelligence},
  volume={37},
  pages={1504--1512},
  year={2023},
  doi={10.1609/aaai.v37i2.25236}
}

@inproceedings{zhou2023tempbalance,
  title={Temperature Balancing, Layer-wise Weight Analysis, and Neural Network Training},
  author={Zhou, Yefan and Pang, Tianyu and Liu, Keqin and Martin, Charles H. and Mahoney, Michael W. and Yang, Yaoqing},
  booktitle={Advances in Neural Information Processing Systems},
  volume={36},
  year={2023}
}

@inproceedings{rajbhandari2020zero,
  title={{ZeRO}: Memory Optimizations Toward Training Trillion Parameter Models},
  author={Rajbhandari, Samyam and Rasley, Jeff and Ruwase, Olatunji and He, Yuxiong},
  booktitle={SC20: International Conference for High Performance Computing, Networking, Storage and Analysis},
  pages={1--16},
  year={2020},
  doi={10.1109/SC41405.2020.00024}
}

@article{zhao2023fsdp,
  title={{PyTorch FSDP}: Experiences on Scaling Fully Sharded Data Parallel},
  author={Zhao, Yanli and Gu, Andrew and Varma, Rohan and Luo, Liang and Huang, Chien-Chin and Xu, Min and Wright, Less and Shojanazeri, Hamid and Ott, Myle and Shleifer, Sam and Desmaison, Alban and Balioglu, Can and Damania, Pritam and Nguyen, Bernard and Chauhan, Geeta and Hao, Yuchen and Mathews, Ajit and Li, Shen},
  journal={Proceedings of the VLDB Endowment},
  volume={16},
  number={12},
  pages={3848--3860},
  year={2023},
  doi={10.14778/3611540.3611569}
}

@article{ross2014mi,
  title={Mutual Information between Discrete and Continuous Data Sets},
  author={Ross, Brian C.},
  journal={PLoS ONE},
  volume={9},
  number={2},
  pages={e87357},
  year={2014},
  doi={10.1371/journal.pone.0087357}
}

@inproceedings{yao2026hclff,
  title={{HCL-FF}: Hierarchical and Contrastive Learning for Forward-Forward Algorithm},
  author={Yao, Jie-En and Chen, Hong-En and Kuo, C.-C. Jay},
  booktitle={Proceedings of the IEEE/CVF Conference on Computer Vision and Pattern Recognition (CVPR)},
  pages={27429--27438},
  year={2026}
}

@article{jiang2026bicovg,
  title={Covariance-Aware Goodness for Scalable Forward-Forward Learning},
  author={Jiang, Xiaoyi and Al-Hashimi, Bashir M. and Xu, Kai},
  journal={arXiv preprint arXiv:2605.04346},
  year={2026}
}

@article{ghader2026forwardonly,
  title={Forward-Only Convolutional Neural Networks with Learnable Channel--Class Assignment},
  author={Ghader, Mohammadnavid and Kheradpisheh, Saeed Reza and Farahani, Bahar and Fazlali, Mahmood},
  journal={arXiv preprint arXiv:2606.09928},
  year={2026}
}

@techreport{krizhevsky2009learning,
  title={Learning Multiple Layers of Features from Tiny Images},
  author={Krizhevsky, Alex},
  institution={University of Toronto},
  year={2009}
}

@article{xiao2017fashion,
  title={Fashion-{MNIST}: a Novel Image Dataset for Benchmarking Machine Learning Algorithms},
  author={Xiao, Han and Rasul, Kashif and Vollgraf, Roland},
  journal={arXiv preprint arXiv:1708.07747},
  year={2017}
}

@inproceedings{coates2011analysis,
  title={An Analysis of Single-Layer Networks in Unsupervised Feature Learning},
  author={Coates, Adam and Ng, Andrew and Lee, Honglak},
  booktitle={Proceedings of the Fourteenth International Conference on Artificial Intelligence and Statistics},
  series={Proceedings of Machine Learning Research},
  volume={15},
  pages={215--223},
  year={2011}
}

@techreport{le2015tiny,
  title={Tiny {ImageNet} Visual Recognition Challenge},
  author={Le, Ya and Yang, Xuan},
  institution={Stanford University, CS 231N},
  year={2015}
}

@inproceedings{deng2009imagenet,
  title={{ImageNet}: A Large-Scale Hierarchical Image Database},
  author={Deng, Jia and Dong, Wei and Socher, Richard and Li, Li-Jia and Li, Kai and Fei-Fei, Li},
  booktitle={2009 IEEE Conference on Computer Vision and Pattern Recognition},
  pages={248--255},
  year={2009},
  doi={10.1109/CVPR.2009.5206848}
}

@article{yang2023medmnist,
  title={{MedMNIST} v2 -- A Large-Scale Lightweight Benchmark for {2D} and {3D} Biomedical Image Classification},
  author={Yang, Jiancheng and Shi, Rui and Wei, Donglai and Liu, Zequan and Zhao, Lin and Ke, Bilian and Pfister, Hanspeter and Ni, Bingbing},
  journal={Scientific Data},
  volume={10},
  pages={41},
  year={2023},
  doi={10.1038/s41597-022-01721-8}
}

@article{kather2019predicting,
  title={Predicting Survival from Colorectal Cancer Histology Slides Using Deep Learning: A Retrospective Multicenter Study},
  author={Kather, Jakob Nikolas and Krisam, Johannes and Charoentong, Pornpimol and Luedde, Tom and Herpel, Esther and Weis, Cleo-Aron and Gaiser, Timo and Marx, Alexander and Valous, Nektarios A. and Ferber, Dyke and Jansen, Lina and Reyes-Aldasoro, Constantino Carlos and Z{\"o}rnig, Inka and J{\"a}ger, Dirk and Brenner, Hermann and Chang-Claude, Jenny and Hoffmeister, Michael and Halama, Niels},
  journal={PLOS Medicine},
  volume={16},
  pages={e1002730},
  year={2019},
  doi={10.1371/journal.pmed.1002730}
}

@article{tschandl2018ham10000,
  title={The {HAM10000} Dataset, a Large Collection of Multi-Source Dermatoscopic Images of Common Pigmented Skin Lesions},
  author={Tschandl, Philipp and Rosendahl, Cliff and Kittler, Harald},
  journal={Scientific Data},
  volume={5},
  pages={180161},
  year={2018},
  doi={10.1038/sdata.2018.161}
}

@article{acevedo2020dataset,
  title={A Dataset of Microscopic Peripheral Blood Cell Images for Development of Automatic Recognition Systems},
  author={Acevedo, Andrea and Merino, Anna and Alf{\'e}rez, Santiago and Molina, {\'A}ngel and Bold{\'u}, Laura and Rodellar, Jos{\'e}},
  journal={Data in Brief},
  volume={30},
  pages={105474},
  year={2020},
  doi={10.1016/j.dib.2020.105474}
}

@inproceedings{loshchilov2019adamw,
  title={Decoupled Weight Decay Regularization},
  author={Loshchilov, Ilya and Hutter, Frank},
  booktitle={International Conference on Learning Representations},
  year={2019}
}

@inproceedings{loshchilov2017sgdr,
  title={{SGDR}: Stochastic Gradient Descent with Warm Restarts},
  author={Loshchilov, Ilya and Hutter, Frank},
  booktitle={International Conference on Learning Representations},
  year={2017}
}

@article{devries2017cutout,
  title={Improved Regularization of Convolutional Neural Networks with Cutout},
  author={DeVries, Terrance and Taylor, Graham W.},
  journal={arXiv preprint arXiv:1708.04552},
  year={2017}
}

@inproceedings{szegedy2016rethinking,
  title={Rethinking the Inception Architecture for Computer Vision},
  author={Szegedy, Christian and Vanhoucke, Vincent and Ioffe, Sergey and Shlens, Jon and Wojna, Zbigniew},
  booktitle={2016 IEEE Conference on Computer Vision and Pattern Recognition (CVPR)},
  pages={2818--2826},
  year={2016},
  doi={10.1109/CVPR.2016.308}
}

@inproceedings{zhang2019rmsnorm,
  title={Root Mean Square Layer Normalization},
  author={Zhang, Biao and Sennrich, Rico},
  booktitle={Advances in Neural Information Processing Systems},
  volume={32},
  year={2019}
}

@inproceedings{simonyan2015vgg,
  title={Very Deep Convolutional Networks for Large-Scale Image Recognition},
  author={Simonyan, Karen and Zisserman, Andrew},
  booktitle={International Conference on Learning Representations},
  year={2015}
}

@article{chen2016checkpoint,
  title={Training Deep Nets with Sublinear Memory Cost},
  author={Chen, Tianqi and Xu, Bing and Zhang, Chiyuan and Guestrin, Carlos},
  journal={arXiv preprint arXiv:1604.06174},
  year={2016}
}

@inproceedings{paszke2019pytorch,
  title={{PyTorch}: An Imperative Style, High-Performance Deep Learning Library},
  author={Paszke, Adam and Gross, Sam and Massa, Francisco and Lerer, Adam and Bradbury, James and Chanan, Gregory and Killeen, Trevor and Lin, Zeming and Gimelshein, Natalia and Antiga, Luca and Desmaison, Alban and K{\"o}pf, Andreas and Yang, Edward and DeVito, Zach and Raison, Martin and Tejani, Alykhan and Chilamkurthy, Sasank and Steiner, Benoit and Fang, Lu and Bai, Junjie and Chintala, Soumith},
  booktitle={Advances in Neural Information Processing Systems},
  volume={32},
  pages={8024--8035},
  year={2019}
}

@article{pedregosa2011sklearn,
  title={Scikit-learn: Machine Learning in {P}ython},
  author={Pedregosa, Fabian and Varoquaux, Ga{\"e}l and Gramfort, Alexandre and Michel, Vincent and Thirion, Bertrand and Grisel, Olivier and Blondel, Mathieu and Prettenhofer, Peter and Weiss, Ron and Dubourg, Vincent and Vanderplas, Jake and Passos, Alexandre and Cournapeau, David and Brucher, Matthieu and Perrot, Matthieu and Duchesnay, {\'E}douard},
  journal={Journal of Machine Learning Research},
  volume={12},
  pages={2825--2830},
  year={2011}
}

@book{cover2006elements,
  title={Elements of Information Theory},
  author={Cover, Thomas M. and Thomas, Joy A.},
  edition={2nd},
  publisher={Wiley-Interscience},
  year={2006}
}

@inproceedings{kingma2015adam,
  title={Adam: A Method for Stochastic Optimization},
  author={Kingma, Diederik P. and Ba, Jimmy},
  booktitle={International Conference on Learning Representations},
  year={2015}
}

@inproceedings{tian2020cmc,
  title={Contrastive Multiview Coding},
  author={Tian, Yonglong and Krishnan, Dilip and Isola, Phillip},
  booktitle={Computer Vision -- ECCV 2020},
  series={Lecture Notes in Computer Science},
  pages={776--794},
  publisher={Springer},
  year={2020},
  doi={10.1007/978-3-030-58621-8_45}
}

\appendix

\section{Formal DTG-FF Description}\label{app:method}

This appendix formalizes DTG-FF components referenced in Sec.~\ref{sec:method}. Subsections cover (A.1) the unified temperature-scaled goodness form and the two loss instances, (A.2) MLP-specific additions (warm-start and label overlay), (A.3) the locality taxonomy including MLP's hybrid classifier path, (A.4) the temperature parametrization, and (A.5) random-projection scaling and the gradient-magnitude bound.

\subsection{Unified Temperature-Scaled Local Goodness and Loss Instances}\label{app:unified}

Let layer $l$ produce post-ReLU activation $\mathbf{z}_l(x) = \mathrm{ReLU}(W_l \hat{\mathbf{x}}_{l-1})$, where $\hat{\mathbf{x}}_{l-1}$ is the detached input from layer $l\!-\!1$. Define a nonnegative goodness representation $\mathbf{u}_l(x) = \phi_l(\mathbf{z}_l(x)) \in \mathbb{R}^{m_l}_{\geq 0}$ and a layer-local learnable temperature $T_l>0$ (Sec.~\ref{sec:dtgmech}). The temperature-scaled goodness is $\tilde{\mathbf{u}}_l = \mathbf{u}_l / T_l$. DTG-FF instances share the form
\begin{equation}
	\mathcal L_l^{\mathrm{FF}}(x,\cdot) = \ell_l\!\big(\psi_l(\tilde{\mathbf{u}}_l),\,\cdot\big),\qquad
	\nabla_{\mathbf{u}_l}\mathcal L_l^{\mathrm{FF}} = T_l^{-1}\, J_{\psi_l}(\tilde{\mathbf{u}}_l)^{\!\top}\,\nabla_{\psi}\ell_l,
	\label{eq:app-unified}
\end{equation}
so temperature modulates the magnitude of the local feature-learning signal in every instance, without altering $I(\mathbf{u}_l;Y)$ under a fixed-batch conditioning.

\paragraph{MLP instance (scalar goodness, margin loss).}
With $m_l\!=\!1$ and $u_l = \tfrac{1}{d_l}\|\mathbf{z}_l\|_2^2$, let $s\!\in\!\{+1,-1\}$ encode positive/negative supervision (Hinton FF convention \citep{hinton2022forward}):
\begin{equation}
	\psi_l^{\mathrm{MLP}}(\tilde u_l) = \tilde u_l - \theta_l,\qquad
	\ell_l^{\mathrm{MLP}}(\psi,s) = \log\!\big(1+\exp(-s\psi)\big).
\end{equation}

\paragraph{CNN instance ($K$-way local cross-entropy).}
With $m_l \gg 1$ and $\mathbf{u}_l = \mathrm{flatten}(\mathrm{AdaptiveAvgPool2d}(\mathbf{z}_l^2,(h_l,w_l)))$:
\begin{equation}
	\psi_l^{\mathrm{CNN}}(\tilde{\mathbf{u}}_l) = \tilde{\mathbf{u}}_l^{\!\top} R_l + \mathbf{b}_l,\qquad
	\ell_l^{\mathrm{CNN}}(\boldsymbol\psi,y) = \mathrm{CE}(\boldsymbol\psi,y),
\end{equation}
with fixed (non-learned) $R_l \in \mathbb{R}^{m_l\times K}$, $R_{l,ij}\!\sim\!\mathcal N(0,1/K)$, and $\mathbf{b}_l\!\in\!\mathbb{R}^K$, $b_{l,k}\!\sim\!\mathcal N(0,1/K)$, registered as buffers.

\paragraph{Why two loss families.}
The objective switch follows convention (\citet{hinton2022forward} for MLP; \citet{papachristodoulou2024cwcomp,gong2026asge} for FF-CNN), supported by a dimensionality argument: when $m_l\!=\!1$, the linear readout $a_k = r_{lk}\,g_l/T_l + b_{lk}$ from a scalar to $K$ logits has rank 1 by construction (image is 1-dimensional), so softmax realizes only a 1-parameter family of class distributions and margin supervision is the natural choice. When $m_l\!\gg\!1$, $K$-way CE is well-defined and we adopt it. We do not claim the dimensionality argument uniquely determines the loss family, only that it explains why the two instantiations differ.

\subsection{MLP-Specific Additions: Warm-Start and Label Overlay}\label{app:warmstart}

The MLP loss in \texttt{main.py} adds a warm-start initialization and the standard Hinton label-overlay procedure on top of the unified form in Eq.~\eqref{eq:app-unified}.

\paragraph{Full MLP per-layer loss.}
\begin{equation}
	\mathcal L_l^{\mathrm{MLP,full}} = -\log\sigma(g_l^{+}/T_l - \theta_l) - \log\sigma(\theta_l - g_l^{-}/T_l),
\end{equation}
where $\theta_l = \mathrm{softplus}(\theta_{0,l})$ is a learnable scalar margin per layer.

\paragraph{Warm-start initialization.}
On the first forward pass, $\theta_{0,l}$ is initialized from the measured average goodness $\bar g_l^{(0)}$: if $\bar g_l^{(0)}\!>\!20$, $\theta_{0,l}\!\leftarrow\!\bar g_l^{(0)}$ (outside the softplus linear regime); otherwise $\theta_{0,l}\!\leftarrow\!\log(\exp(\bar g_l^{(0)})-1)$ (inverse softplus). This places the initial decision boundary at layer-specific activation scales rather than at $\theta\!=\!0$.

\paragraph{Label overlay.}
In the MLP, the input to each FF forward pass is not the raw image $x$ but $x$ with its first $K$ entries overwritten by $\lambda \mathbf{e}_y$ (one-hot target scaled by $\lambda\!=\!1.0$ default). The positive pass uses $y^{+}\!=\!y$ (true label) and the negative pass uses $y^{-}\!=\!\mathrm{random}(\{0,\dots,K\!-\!1\}\setminus\{y\})$. The overlay is the standard FF procedure \citep{hinton2022forward} and is omitted from the CNN instance, which uses CE on all $K$ labels and therefore does not require a separate negative construction.

\subsection{Locality Taxonomy and MLP Hybrid Classifier}\label{app:locality}

We distinguish three locality properties:
\begin{description}
	\item[Inter-layer gradient locality.] Updates to $W_l$ require no gradient propagated through layers $j\!\neq\!l$ in the relevant loss.
	\item[Batch locality.] Updates may use statistics aggregated across samples within the same minibatch (as in BatchNorm).
	\item[Sample locality.] Updates for sample $x_b$ use only statistics derivable from $x_b$ alone.
\end{description}
The FF (goodness) loss in both MLP and CNN DTG-FF satisfies \emph{inter-layer gradient locality} (enforced by \texttt{detach} of $\mathbf{z}_l$ before passing to layer $l\!+\!1$) and \emph{sample locality} in the FF path: since $T_l$ and $\theta_l$ are learnable parameters of layer $l$ rather than batch statistics, the per-layer loss for sample $x_b$ depends only on $x_b$ and layer-$l$ parameters.

\paragraph{Hybrid classifier head (MLP).}
The MLP pipeline combines the FF per-layer loss with a global CE over a classifier built from the \emph{same} layer-wise linear weights: for each layer $i<L\!-\!1$, the classifier input is $\mathrm{BN}_i(W_i z_{i-1})$ (no \texttt{detach}), concatenated across layers and fed into a linear readout. The joint objective is
\begin{equation}
	\mathcal L^{\mathrm{MLP,total}} = \lambda_{\mathrm{ff}}\sum_{l=0}^{L-1} \mathcal L_l^{\mathrm{MLP,full}} \;+\; \mathcal L^{\mathrm{CE}}_{\mathrm{cls}},
\end{equation}
with default $\lambda_{\mathrm{ff}}\!=\!0.1$. Consequently, the classifier loss backpropagates end-to-end through the layer weights, and the MLP variant is a \emph{hybrid} FF+BP system (consistent with the auxiliary-classifier tradition \citep{belilovsky2019greedy}) rather than a strictly inter-layer-local training procedure. We state this explicitly to avoid readers inferring pure biological plausibility.

\paragraph{CNN pipeline.}
The CNN classifier path is separated from conv training by \texttt{torch.no\_grad()} wrappers during feature extraction (see Sec.~\ref{sec:dtgcnn}); the conv backbone receives gradients only from its own per-layer FF loss. The CNN is therefore strictly inter-layer-gradient-local in the conv backbone.

\paragraph{MLP vs.\ CNN classifier index sets differ.}
The MLP hybrid classifier above uses layers $i\!<\!L\!-\!1$ (i.e., excludes the \emph{last} hidden layer). The CNN concat classifier (Sec.~\ref{sec:dtgcnn}) uses layers $1,\ldots,L\!-\!1$ in 0-based indexing (i.e., excludes the \emph{first} conv layer). The two variants therefore drop opposite ends; this is a design choice of our implementations rather than a convention taken from prior work, and both choices are matched by the implementations and are not load-bearing claims.

\subsection{Temperature Parametrization}\label{app:ema}

Each layer's temperature is parametrized as $T_l = T_{\min} + (T_{\max} - T_{\min})\,\sigma(\alpha_l)$, where $\alpha_l\!\in\!\mathbb{R}$ is a single learnable scalar per layer (initialized to $0$, giving $T_l = (T_{\min} + T_{\max})/2$ at the start of training) and $\sigma$ is the logistic sigmoid. The bound $T_l\!\in\![T_{\min}, T_{\max}]$ with $T_{\min}\!=\!0.5$, $T_{\max}\!=\!2.0$ for CNN and $T_{\min}\!=\!0.1$, $T_{\max}\!=\!2.0$ for MLP prevents $T_l$ from collapsing to $0$ or diverging. $T_l$ is updated alongside $W_l$ via gradient descent on the layer-local loss (Eq.~\eqref{eq:app-unified}).

At inference on the CNN, the per-layer logit $\psi_l^{\mathrm{CNN}}(\mathbf{u}_l/T_l) = (\mathbf{u}_l/T_l)^{\!\top}R_l + \mathbf{b}_l$ retains the temperature scaling (since $T_l$ is a fixed learned parameter). The MLP goodness loss is only evaluated during training; at test time the MLP uses its classifier head without invoking the margin objective.

\subsection{Random Projection Scale and Gradient-Magnitude Bound}\label{app:rscale}

For the CNN instance, $\hat{\mathbf{y}}_l = (\mathbf{u}_l/T_l)^{\!\top}R_l + \mathbf{b}_l$ with fixed $R_l\!\in\!\mathbb R^{m_l\times K}$, $R_{l,ij}\!\sim\!\mathcal N(0,1/K)$. The CE gradient w.r.t.\ goodness is
\begin{equation}
	\nabla_{\mathbf{u}_l}\mathcal L_l^{\mathrm{CNN}} = T_l^{-1}\,R_l\,(\mathbf{p}-\mathbf{e}_y),
	\quad \mathbf{p}=\mathrm{softmax}(\hat{\mathbf{y}}_l),
	\quad \|\mathbf{p}-\mathbf{e}_y\|_2 \le \sqrt{2},
\end{equation}
so
\begin{equation}
	\|\nabla_{\mathbf{u}_l}\mathcal L_l^{\mathrm{CNN}}\|_2 \;\le\; \frac{\sqrt{2}}{T_l}\,\|R_l\|_{\mathrm{op}},
	\label{eq:app-grad-bound}
\end{equation}
where $\|R_l\|_{\mathrm{op}}$ is the spectral norm. For class columns $r_k\!\in\!\mathbb R^{m_l}$, $\mathbb E\|r_k\|_2^2 = m_l/K$; with typical FF-CNN sizes $m_l\!\in\!\{512,1024,2048\}$ and $K\!=\!10$, typical column norms are $O(\sqrt{m_l/K})\!\sim\!7\!-\!15$.

\paragraph{Temperature ranges differ by architecture.}
For CNN, $T_l\!\in\![0.5,2.0]$, so the DTG prefactor $1/T_l$ varies by at most $T_{\max}/T_{\min}\!=\!4$. For MLP, $T_l\!\in\![0.1,2.0]$, with a ratio of $20$; the stronger MLP modulation is consistent with the coarser scalar-goodness signal needing larger relative gradient-scale adjustment.

\paragraph{Scale-confound ablation.}
To check whether DTG's gain is merely compensating for per-column norm variance in the Gaussian-sampled $R_l$, we run a 4-cell ablation on CIFAR-10 (VGG8, 400 epochs, seed~42) crossing $\{T\!=\!1, \mathrm{DTG}\} \times \{\text{unnormalized}~R_l, \text{unit-column-norm}~\tilde R_l\}$:
\begin{center}
	\small
	\setlength{\tabcolsep}{8pt}
	\begin{tabular}{@{}lccc@{}}
		\toprule
		Temperature                  & unnormalized $R_l$ & normalized $\tilde R_l$ & $\Delta$ (unnorm $-$ norm) \\
		\midrule
		$T\!=\!1$ (no DTG)           & 90.61\%            & 89.71\%                 & $+0.90$                    \\
		DTG (learnable $T_l$)        & 91.33\%            & 90.22\%                 & $+1.11$                    \\
		$\Delta$ (DTG $-$ $T\!=\!1$) & $+0.72$            & $+0.51$                 & ---                        \\
		\bottomrule
	\end{tabular}
\end{center}

Two findings. First, DTG's gain over $T\!=\!1$ persists under column-normalized $R_l$ ($+0.51\%$), so DTG is not purely compensating for random-column-norm variation---roughly $0.51/0.72\!\approx\!71\%$ of DTG's advantage is independent of $R_l$ scale. Second, normalizing $R_l$ columns hurts both $T\!=\!1$ and DTG configurations by $\approx\!1$ percentage point, suggesting the default Gaussian column-norm variation acts as a mild implicit regularizer. We therefore do not adopt column normalization in the default implementation, and we report the findings here to pre-empt the scale-compensation concern.

\section{Extended Related Work}\label{app:related}

This appendix expands Sec.~\ref{sec:related} with coverage too long for the 9-page main text.

\paragraph{Forward-Forward variants.}
\citet{lee2023symba} (SymBa) proposed a symmetric contrastive FF loss to improve convergence. \citet{papachristodoulou2024cwcomp} (CwComp) introduced channel-wise competitive learning for FF-CNNs. \citet{chen2025scff} (SCFF, Nature Comm.) construct positive and negative examples self-contrastively from the data, train without labels and evaluate with a linear readout; they also extend FF to recurrent networks. \citet{sun2025deeperforward} (DeeperForward, ICLR 2025) combine layer normalization with mean goodness and train all layers concurrently with a pipelined scheme, reaching 88.72\% on CIFAR-10 with 1000 epochs and stronger augmentation. \citet{gong2026asge} (ASGE) supervise adaptive spatial goodness encodings layer-wise and fuse concatenated GAP features, reaching 90.62\%---the prior FF-family best at the time of our submission---and report FF training on ImageNet-1k. \citet{ororbia2023pff} (PFF) combined FF with predictive coding, re-introducing partial inter-layer coordination.

\paragraph{Spatial goodness attribution.}
We want to clearly credit prior work: vector-valued goodness with cross-entropy readouts was used in prior FF-CNN work \citep{papachristodoulou2024cwcomp}, and spatial goodness encoding specifically follows ASGE \citep{gong2026asge}. Our Pathway~1 (signal quality) in Sec.~\ref{sec:diag} identifies vector goodness as one of the three DTG-FF components, but the \emph{contribution} of this work is DTG (temperature mechanism), decoupled normalization, and multi-layer integration, not spatial goodness per se.

\paragraph{Biologically plausible BP alternatives.}
Feedback alignment replaces the transpose of forward weights with random feedback \citep{lillicrap2016feedback}; direct feedback alignment projects errors directly to each hidden layer \citep{nokland2016direct,launay2020dfa}. Target propagation computes layer-wise targets via learned inverse mappings \citep{bengio2014target,lee2015target,meulemans2020theoretical}. Equilibrium propagation derives local, contrastive (Hebbian-like) updates from a free and a nudged equilibrium phase of an energy-based network \citep{scellier2017equilibrium}, and later work relaxes its weight-symmetry requirement through Jacobian homeostasis \citep{laborieux2024improving}; \citet{millidge2023backpropagation} showed predictive coding, EqProp, and contrastive Hebbian learning approximate BP in a unified limit. Predictive coding networks provide structured backward signals via prediction errors \citep{rao1999predictive,whittington2017approximation,millidge2022predictive,salvatori2024stable}. PEPITA modulates the input with the output error \citep{dellaferrera2022pepita}, and forward-gradient methods estimate gradients from activity perturbations with local losses \citep{ren2023forward}. An \emph{informal} ordering by retained backward-signal richness (BP $>$ FA/DFA $>$ TP $>$ PC $>$ FF) helps position FF as the most constrained end of the design space; this ordering is intuitive rather than formally established, since ``retained information'' is not uniformly defined across methods. DTG-FF preserves the zero-backward-gradient property in its FF goodness path.

\paragraph{Auxiliary-classifier heads with local losses.}
Deeply-supervised networks \citep{lee2015dsn}, GoogLeNet \citep{szegedy2015going}, and greedy layer-wise training \citep{belilovsky2019greedy,belilovsky2020decoupled} attach classifiers to intermediate layers. \citet{nokland2019training} used similarity matching and prediction auxiliary losses. These differ from FF in that classifiers' gradients typically backpropagate into convolutional layers (providing richer supervision than scalar goodness). DTG-FF's CNN classifier is \texttt{detach}-separated from the conv backbone; the MLP classifier is a hybrid (App.~\ref{app:locality}).

\paragraph{Information-theoretic perspectives on deep learning.}
\citet{tishby1999information} introduced the information bottleneck; \citet{tishby2015deep} proposed deep networks implicitly optimize it; \citet{shwartzziv2017opening} provided empirical support. \citet{saxe2018information} showed compression depends on activation function---ReLU networks do not compress in the same sense as tanh networks. MI estimation in high dimensions is notoriously hard \citep{belghazi2018mine,poole2019variational,mcallester2020limitations}. Our measurements (App.~\ref{app:mi_methodology}) avoid deep estimator reliance by using scalar histogram and $k$-nearest-neighbour estimates \citep{kraskov2004ksg,ross2014mi} and linear-probe Fano bounds.

\paragraph{Normalization and temperature scaling.}
\citet{santurkar2018does} argued that BN helps optimization by smoothing the loss landscape rather than by reducing internal covariate shift, while mean-field analysis shows that BN itself induces gradient explosion at initialization in deep networks without skip connections \citep{yang2019mean}. \citet{ba2016layer} introduced LayerNorm as a batch-independent alternative. Temperature scaling is prominent in distillation \citep{hinton2015distilling} and calibration \citep{guo2017calibration}; curriculum-scheduled distillation temperatures \citep{li2023curriculum} operate on softmax outputs, and TempBalance \citep{zhou2023tempbalance} sets layer-wise learning-rate ``temperatures'' from heavy-tailed spectral metrics; both operate within BP-trained networks. DTG differs by scaling the local \emph{learning signal} in a layer-local loss, not the softmax output.

\section{Information Diagnostic Methodology}\label{app:mi_methodology}

\subsection{Setup}\label{app:mi}

\begin{figure}[h]
	\centering
	\includegraphics[width=0.6\textwidth]{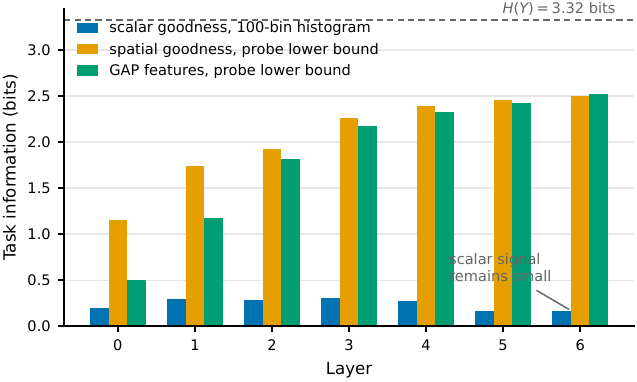}
	\caption{Per-layer task-information diagnostic on a trained DTG-FF VGG8 (CIFAR-10, $91.33\%$). Scalar goodness uses a 100-bin histogram estimate; spatial goodness and GAP report linear-probe Fano lower bounds.}
	\label{fig:mi}
\end{figure}

Measurements use the DTG-FF VGG8 trained on CIFAR-10 (concat-classifier variant, test accuracy $91.33\%$). We extract per-layer activations from the test set ($10{,}000$ images) after the ReLU and before inter-layer normalization. For each layer $l\!\in\!\{0,\dots,6\}$ we compute:
\begin{itemize}
	\tightlist
	\item scalar goodness $g_l = \tfrac{1}{d_l}\|\mathbf{z}_l\|_2^2 \in \mathbb{R}$ per sample,
	\item spatial goodness vector $\mathbf{g}_l^{\mathrm{vec}} = \mathrm{flatten}(\mathrm{AAP}(\mathbf{z}_l^2,(h_l,w_l))) \in \mathbb{R}^{m_l}$,
	\item GAP features $\mathbf{f}_l^{\mathrm{GAP}} = \tfrac{1}{H_l W_l}\sum_{h,w} z_{l,c,h,w} \in \mathbb{R}^{C_l}$.
\end{itemize}

\subsection{Estimators}\label{app:mi_estimators}
\paragraph{Scalar MI.} The per-layer scalar estimate of $I(g_l;Y)$ (table below, Fig.~\ref{fig:mi}) uses a 100-bin histogram. As a sensitivity check we also use bin counts $\{50,200,500\}$ and the $k$-nearest-neighbour estimator of \citet{ross2014mi}, a KSG-style \citep{kraskov2004ksg} extension to continuous-feature/discrete-target MI, via \texttt{sklearn.feature\_selection.mutual\_info\_classif} \citep{pedregosa2011sklearn} with its default $k\!=\!3$ (\texttt{random\_state=42}).

\paragraph{Vector MI lower bound.} For $\mathbf{g}^{\mathrm{vec}}_l$ and $\mathbf{f}^{\mathrm{GAP}}_l$, we train a linear probe (softmax regression) on an 80/20 split of the test set---50 epochs Adam \citep{kingma2015adam}, $\mathrm{lr}\!=\!0.01$, inputs standardized by training-split statistics. Let $\hat Y$ be the probe's prediction on held-out samples with error $P_e = \Pr[\hat Y\!\neq\!Y]$. Since $\hat Y$ is a deterministic function of the layer features (call them $F$), the data-processing inequality gives $I(F;Y) \geq I(\hat Y;Y)$. By Fano's inequality \citep{cover2006elements} with uniform prior,
\begin{equation}
	I(\hat Y;Y) \geq \log_2 K - H_2(P_e) - P_e \log_2(K-1),
\end{equation}
where $H_2$ is binary entropy. This lower bound depends on linear-probe separability and is not representation-intrinsic.

\subsection{Per-layer Measurements}\label{app:observables}
\begin{center}
	\small
	\setlength{\tabcolsep}{5pt}
	\begin{tabular}{@{}crccc@{}}
		\toprule
		Layer & $\dim(\mathbf{g}^{\mathrm{vec}}_l)$ & $I(g;Y)$ (hist.) & $I(\mathbf{g}^{\mathrm{vec}};Y)\!\geq$ & $I(\mathrm{GAP};Y)\!\geq$ \\
		\midrule
		0     & 2048                                & 0.20           & 1.15                                   & 0.50                      \\
		1     & 1024                                & 0.30           & 1.74                                   & 1.17                      \\
		2     & 1024                                & 0.28           & 1.92                                   & 1.82                      \\
		3     & 512                                 & 0.31           & 2.26                                   & 2.17                      \\
		4     & 512                                 & 0.27           & 2.39                                   & 2.33                      \\
		5     & 512                                 & 0.16           & 2.46                                   & 2.42                      \\
		6     & 512                                 & 0.16           & 2.50                                   & 2.52                      \\
		\midrule
		Mean  & ---                                 & 0.24           & 2.06                                   & 1.85                      \\
		\bottomrule
	\end{tabular}
\end{center}

\paragraph{Histogram sensitivity.} Mean $I(g;Y)$ across layers: 0.22 bits (50 bins), 0.24 (100), 0.28 (200), 0.38 (500). Histogram estimates inflate with bin count (finite-sample bias); the $k$-nearest-neighbour estimator ($k\!=\!3$) gives 0.20 bits, consistent with the lower end of the histogram range.

\begin{figure}[t]
	\centering
	\includegraphics[width=0.78\textwidth]{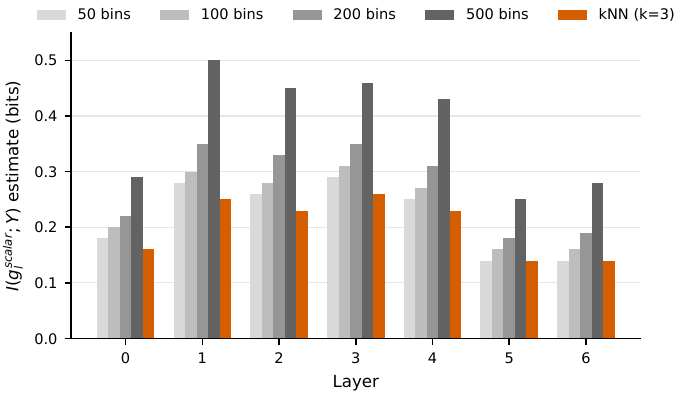}
	\caption{Sensitivity of scalar-goodness mutual-information estimates to estimator choice. Histogram estimates increase with bin count, consistent with finite-sample bias; the $k$-nearest-neighbour estimate remains near the lower end of the histogram range.}
	\label{fig:mi-sensitivity}
\end{figure}

\paragraph{Per-layer observables for Pathway~3.}
Using logit-sum inference on the concat-classifier checkpoint (CIFAR-10, $91.33\%$), per-layer accuracies are $\{43.5,77.6,85.1,88.7,91.2,91.3,91.0\}\%$. Pairwise prediction disagreement averages $25.1\%$ (off-diagonal mean); the oracle-ensemble accuracy (``any layer correct'') is $96.9\%$ against the logit-sum-evaluated ensemble's $91.5\%$ (distinct from the $91.79\%$ headline obtained by training a dedicated logit-sum model from scratch, since here the random-projection heads were never the optimization target), a gap of $5.4$ percentage points. Sec.~\ref{sec:diag} reports these values as observable support for the three-pathway lens.

\paragraph{Limitations of the diagnostic.}
These numbers are architecture- and checkpoint-specific. The scalar estimate is consistent across binning and estimators but is only measured on DTG-FF VGG8 (CIFAR-10). The vector numbers are linear-probe lower bounds: a different probe class could yield a different bound. We do not claim the MI measurements upper-bound achievable FF performance; they are diagnostic of the current trained representations.

\section{Extended Experimental Details}\label{app:extra_experiments}

\subsection{Full Hyperparameters}\label{app:hyper}

\begin{center}
	\small
	\setlength{\tabcolsep}{6pt}
	\begin{tabular}{@{}p{0.32\linewidth}p{0.62\linewidth}@{}}
		\toprule
		Setting                                 & Value                                                                                                                                        \\
		\midrule
		Backbone (32$\times$32, 64$\times$64)   & VGG8: 7 conv layers, $3\!\to\!128\!\to\!256\!\to\!256\!\to\!512\!\to\!512\!\to\!512\!\to\!512$ (input$\to$conv outputs); $2\!\times\!2$ AvgPool after layers 1,3,4,5 \\
		Backbone (224$\times$224)               & VGG11: 8 conv layers, $3\!\to\!64\!\to\!128\!\to\!256\!\to\!256\!\to\!512\!\to\!512\!\to\!512\!\to\!512$                                                                  \\
		FF optimizer (per layer)                & AdamW, $\mathrm{lr}\!=\!2\!\times\!10^{-4}$, weight decay $10^{-3}$, cosine to $10^{-5}$, grad-norm clip $1.0$                                                     \\
		Classifier optimizer                    & AdamW, same hyperparameters (incl.\ clip $1.0$)                                                                                                                  \\
		Temperature range CNN                   & $T_l\!\in\![0.5, 2.0]$ parametrized as $T_l\!=\!T_{\min}\!+\!(T_{\max}\!-\!T_{\min})\sigma(\alpha_l)$, one $\alpha_l$ per layer              \\
		Temperature range MLP                   & $T\!\in\![0.1, 2.0]$                                                                                                                         \\
		Epochs                                  & 400 for 32$\times$32; 200 for Tiny ImageNet (64$\times$64) and ImageNet-100 (224$\times$224)                                                 \\
		Batch size                              & 128 (32$\times$32), 64 (64$\times$64 and 224$\times$224)                                                                                     \\
		Augmentation                            & Random crop (padding 4), flip, color jitter, grayscale ($p\!=\!0.2$), $16\!\times\!16$ Cutout                                                             \\
		Inter-layer dropout / norm              & 0.1 / LayerNorm across channel dim per $(h,w)$ location                                                                                      \\
		Goodness pool size $(h_l, w_l)$         & ASGE-style \citep{gong2026asge} schedule by output channel count: $C_l\!=\!64\!\to\!8\!\times\!8$, $128\!\to\!4\!\times\!4$, $256\!\to\!2\!\times\!2$, $\geq\!512\!\to\!1\!\times\!1$ \\
		Classifier dropout / label smoothing    & 0.2 / 0.1                                                                                                                                    \\
		Loss blending (MLP)                     & $\lambda_{\mathrm{ff}}\!=\!0.1$ (Eq.~9 in App.~\ref{app:locality})                                                                           \\
		\bottomrule
	\end{tabular}
\end{center}

\subsection{Secondary Benchmarks}\label{app:secondary}

DTG-FF's pattern generalizes beyond CIFAR and ImageNet-scale datasets. Table below reports results on 5 additional benchmarks (VGG8, 400 epochs, same hyperparameters unless noted).

\begin{center}
	\small
	\setlength{\tabcolsep}{8pt}
	\begin{tabular}{@{}lcccl@{}}
		\toprule
		Dataset       & Classes & Res. & DTG-FF  & Notes                                                  \\
		\midrule
		Fashion-MNIST & 10      & 28   & 94.67\% & grayscale expanded to 3ch                              \\
		STL-10        & 10      & 96   & 82.85\% & native 96$\times$96, 5k train images                   \\
		PathMNIST     & 9       & 28   & 89.82\% & colon pathology                                        \\
		DermaMNIST    & 7       & 28   & 76.56\% & dermatoscopy (intrinsically hard)                       \\
		BloodMNIST    & 8       & 28   & 96.61\% & blood cell, approaches BP-level                        \\
		\bottomrule
	\end{tabular}
\end{center}

The BloodMNIST result is the standout: DTG-FF approaches BP-level performance on this tractable texture-based task (MedMNIST v2 BP baselines: 95--97\% \citep{yang2023medmnist}). DermaMNIST's lower accuracy reflects dataset difficulty (small training set, high intra-class visual similarity); MedMNIST v2 BP baselines report 73--77\% \citep{yang2023medmnist}. Fashion-MNIST and the MedMNIST datasets (natively 28$\times$28) are upsampled to 32$\times$32; STL-10 is trained at its native 96$\times$96.

\subsection{MLP Ablation (CIFAR-10)}\label{app:mlp}

We report the MLP progressive ablation on CIFAR-10 (4 layers, hidden dim 2048, 100 epochs). This is distinct from the CNN variant used in Sec.~\ref{sec:ablation}.
\begin{center}
	\small
	\setlength{\tabcolsep}{8pt}
	\begin{tabular}{@{}lc@{}}
		\toprule
		Configuration                                       & Accuracy                   \\
		\midrule
		Original FF with BN everywhere                      & $\sim\!10\%$ (near chance) \\
		Pure FF (no classifier)                             & 16--17\%                   \\
		DTG-FF 4L (no classifier BN)                        & 46.00\%                    \\
		\ \ + multi-layer concat classifier                 & 52.51\%                    \\
		\ \ + $\lambda_{\mathrm{ff}}\!=\!0.1$ loss blending & 53.32\%                    \\
		\ \ + LayerNorm before classifier                   & 55.43\%                    \\
		\ \ + per-layer BN in classifier path               & \textbf{63.72\%}           \\
		\bottomrule
	\end{tabular}
\end{center}
Per-layer BN in the classifier path is the largest single-step MLP improvement ($+8.29\%$), consistent with the decoupled-normalization principle: BN helps the classifier path on detached features while being destructive in the FF goodness path. The final 63.72\% exceeds the original FF's $\sim\!59\%$ on CIFAR-10 with locally connected networks \citep{hinton2022forward}; note that our MLP is a hybrid FF+BP system (App.~\ref{app:locality}).

\subsection{Extended CNN Ablation Analysis}\label{app:ablation}

\begin{table}[h]
	\centering
	\small
	\setlength{\tabcolsep}{5pt}
	\renewcommand{\arraystretch}{0.95}
	\caption{Full CIFAR-10 ablation rows including Cutout, RMSNorm, and regularization variants. Headline rows in Sec.~\ref{sec:ablation}.}\label{tab:ablation_full}
	\begin{tabular}{@{}lcr@{}}
		\toprule
		Configuration                                         & Accuracy & $\Delta$ \\
		\midrule
		DTG-FF (concat) baseline                              & 91.33\%  & ---      \\
		\ \ replace LayerNorm with RMSNorm$^{\dagger}$        & 90.25\%  & $-0.18$  \\
		\ \ $-$ Cutout augmentation                           & 91.27\%  & $-0.06$  \\
		\ \ $-$ regularization (label smoothing + cls dropout) & 91.50\% & $+0.17$  \\
		\bottomrule
	\end{tabular}\\
	\footnotesize{$^{\dagger}$RMSNorm delta measured at 200 epochs, relative to LayerNorm at 200 epochs ($90.43\%$); other rows at 400 epochs.}
\end{table}

We note two complementary observations:

\paragraph{Cosine annealing as implicit regularizer.}
With a cosine learning-rate schedule from $2\!\times\!10^{-4}$ to $10^{-5}$ over 400 epochs, explicit regularization (label smoothing 0.1 + classifier dropout 0.2) has a small effect on test accuracy ($+0.17\%$ when removed) but dramatically changes training classifier loss ($0.25$ without regularization vs. $0.87$ with). This suggests the cosine schedule provides substantial implicit regularization on CIFAR-10. Whether this holds on larger datasets (Tiny ImageNet, ImageNet-100) where overfitting pressure is higher requires separate study.

\paragraph{Label smoothing interaction.}
Ablating label smoothing (classifier CE loss) yields $+0.17\%$ on CIFAR-10 at 400 epochs, i.e., \emph{removing} it slightly improves accuracy. We interpret this as suggestive that label smoothing and DTG's temperature scaling overlap functionally: both soften the effective decision boundary during training. Under strong cosine annealing, the implicit regularization already suffices, and label smoothing becomes slightly redundant. The effect size is within the expected single-seed noise band; whether label smoothing consistently harms DTG-FF in this regime requires multi-seed follow-up.

\subsection{\texorpdfstring{Depth Scaling: VGG8 vs VGG11 on 32$\times$32 Inputs}{Depth Scaling: VGG8 vs VGG11 on 32x32 Inputs}}\label{app:depth}

We probe DTG-FF's depth-scaling behavior by training the deeper VGG11 backbone (8 conv layers, $9.3$M params, vs.\ VGG8's $9.2$M and 7 conv layers, CIFAR-10 heads) on CIFAR-10/100 under identical hyperparameters (AdamW lr $2\!\times\!10^{-4}$, cosine to $10^{-5}$, 400 epochs, batch 128, same augmentation). Results:

\begin{center}
	\small
	\setlength{\tabcolsep}{8pt}
	\begin{tabular}{@{}lcccc@{}}
		\toprule
		Dataset   & $K$ & DTG-FF VGG8 & DTG-FF VGG11 & $\Delta$          \\
		\midrule
		CIFAR-10  & 10  & 91.33\%     & 84.36\%      & $-6.97$           \\
		CIFAR-100 & 100 & 67.28\%     & 54.79\%      & $\mathbf{-12.49}$ \\
		\bottomrule
	\end{tabular}
\end{center}

The VGG11 accuracy drop is $-6.97$ pp on CIFAR-10 ($K\!=\!10$) and $-12.49$ pp on CIFAR-100 ($K\!=\!100$); the cost nearly doubles as class count increases. \textbf{Width-vs-depth confound.} VGG11 in our setup starts with 64 channels (vs.\ VGG8's 128) and applies $2\!\times\!2$ pooling at the first layer, while VGG8 retains full resolution through layer~1; VGG11 thus differs from VGG8 in depth (8 vs.\ 7 conv layers), early-channel width (64 vs.\ 128), and downsampling schedule. The accuracy drop reported here therefore bundles depth and width effects, and we cannot strictly attribute the cost to depth alone. A depth-only ablation (matched widths, varying number of layers) is left to future work. Combined with the synthetic $K$-scaling observed in Sec.~\ref{sec:synthetic}, we observe that local-learning cost grows with task complexity and with deeper/narrower architectures together; this is consistent with prior FF-family reports that depth alone does not monotonically improve FF accuracy \citep{sun2025deeperforward}. Depth-specific calibration (layer-wise learning rate scaling, warmup, or architectural choices such as the residual design of DeeperForward \citep{sun2025deeperforward}) may mitigate this cost but is left to future work.

\paragraph{Why does VGG11 recover on ImageNet-100?}
The same VGG11 backbone reaches $49.4\%$ on ImageNet-100 at $224\!\times\!224$ (Sec.~\ref{sec:sota}). Although the absolute accuracy is lower than VGG11's $54.79\%$ on CIFAR-100, ImageNet-100 has $100$ classes at substantially higher visual complexity, and each $224\!\times\!224$ image carries far more per-sample information than a $32\!\times\!32$ CIFAR image (representing dozens to hundreds of ``effective receptive fields'' worth of content). We hypothesize that higher input signal per sample partially compensates the depth cost that manifests acutely on $32\!\times\!32$ inputs. Under this view, VGG11 is a defensible choice for high-resolution tasks even though it underperforms VGG8 on $32\!\times\!32$.

\paragraph{Hyperparameter comparison with ASGE.}
ASGE's CIFAR recipe \citep{gong2026asge} matches ours in optimizer (AdamW), learning rate ($2\!\times\!10^{-4}$ cosine to $10^{-5}$), weight decay ($10^{-3}$), batch size (128) and 400 epochs. It differs in using a 45k/5k train/validation split with validation-based checkpoint selection and averaging three runs, whereas our numbers are single-seed with test-set checkpoint selection (Sec.~\ref{sec:setup}). The $6.26$ pp CIFAR-10 difference between ASGE-VGG11 ($90.62\%$) and our VGG11 ($84.36\%$) therefore cannot be explained by these optimization hyperparameters; it more plausibly reflects differences in goodness encoding, readout and architecture details, which we did not isolate. VGG8 remains our operating point for headline results.

\subsection{Within-Dataset $K$ Probe: Full Table}\label{app:granularity}

\begin{center}
	\small
	\setlength{\tabcolsep}{6pt}
	\begin{tabular}{@{}lrcccc@{}}
		\toprule
		Granularity & $K$   & DTG-FF (concat) & BP-DeepSup & FF--BP gap \\
		\midrule
		Coarse      & $20$  & $66.14\%$       & $74.13\%$  & $7.99$ pp  \\
		Fine        & $100$ & $51.28\%$       & $62.63\%$  & $11.35$ pp \\
		\midrule
		\multicolumn{4}{r}{\emph{Within-dataset gap differential:}} & $3.36$ pp \\
		\bottomrule
	\end{tabular}
\end{center}

VGG8, batch $128$, AdamW lr $2\!\times\!10^{-4}$, cosine schedule to $10^{-5}$, $50$ epochs, single seed (114514). Identical augmentation across granularities; only the label mapping changes. Both granularities are non-converged at $50$ epochs (the $400$-epoch CIFAR-100 fine result in Sec.~\ref{sec:sota} is a $5.93$ pp gap, smaller than the $11.35$ pp at $50$ epochs because BP and DTG-FF benefit from extended training at different rates). The relative ordering---coarse gap $<$ fine gap---is the load-bearing observation.

\subsection{Memory-Feasibility Frontier: Detailed Measurements and Fair BP Baselines}\label{app:memory}

\paragraph{Hardware and protocol.}
NVIDIA RTX 4060 Laptop ($8.0$ GB VRAM), PyTorch \citep{paszke2019pytorch} $2.9.1$ + CUDA $13.0$, CPU pinned dataloader off (synthetic random tensors). For each (method, batch, arch, input size) cell we run $5$ warmup steps then time $20$ measured steps under \texttt{torch.cuda.synchronize}; peak memory uses \texttt{torch.cuda.max\_memory\_allocated} after \texttt{reset\_peak\_memory\_stats}. We report peak allocated (not reserved) memory and per-step wallclock.

\paragraph{Naive vs.\ pipelined DTG-FF.}
The naive implementation in our initial release runs \texttt{forward\_ff} (with grad) and \texttt{forward\_features} (\texttt{torch.no\_grad}) sequentially, recomputing the convolution twice. The pipelined schedule integrates the two into a single \texttt{train\_step}: forward $\to$ backward $\to$ optimizer step $\to$ derive next-layer input from \texttt{z\_act.detach()} \emph{after} the autograd graph has been freed. The order matters: deriving the next-layer input before the local backward causes the propagation tensors to coexist with the autograd state and increases peak memory. The pipelined version is mathematically identical for each layer's FF loss; only the propagation semantics differ by one optimizer step (the propagated input reflects pre-step weights of the just-trained layer rather than post-step weights), a perturbation of order $\mathrm{lr}$ on activations. We verified empirically that this does not affect convergence trajectory on CIFAR-10 (5-epoch sanity train: $59.84\%$ test, normal trajectory).

\paragraph{Depth scaling at $32\!\times\!32$ batch $128$.}
\begin{center}
	\small
	\setlength{\tabcolsep}{6pt}
	\begin{tabular}{@{}lcccc@{}}
		\toprule
		Arch  & DTG-FF concat (peak / ms) & BP-VGG single (peak / ms) & Memory $\Delta$ & Speed $\Delta$ \\
		\midrule
		VGG8  & $0.846$ GB / $94$ ms  & $0.870$ GB / $83$ ms  & $-2.7\%$  & DTG $+13\%$ slower \\
		VGG11 & $0.320$ GB / $52$ ms  & $0.333$ GB / $20$ ms  & $-4.1\%$  & DTG $\times 2.6$ slower \\
		VGG13 & $0.355$ GB / $64$ ms  & $0.430$ GB / $28$ ms  & $-17.4\%$ & DTG $\times 2.3$ slower \\
		\bottomrule
	\end{tabular}
\end{center}
At small input size, BP runs as a single fused forward and backward and amortizes per-iteration overhead across all layers; DTG-FF performs $L$ separate backward calls and thus pays per-layer kernel-launch overhead. The per-architecture comparisons here are confounded by width and downsampling differences across our VGG variants---VGG11 starts with narrower channels (64) and pools earlier than VGG8 (128, no first-layer pool), which dominates the per-architecture peak-memory difference at $32\!\times\!32$ and explains why VGG11's footprint sits below VGG8's despite having more conv layers. The structural $\mathcal{O}(L)$-vs-$\mathcal{O}(1)$ activation-memory contrast between BP and pipelined DTG-FF only manifests cleanly when input size and per-layer activation volume are large; at $224\!\times\!224$ (Sec.~\ref{sec:systems}) DTG-FF's bound becomes the binding factor, while at $32\!\times\!32$ both methods sit in a regime where width allocation matters more than depth.

\paragraph{Throughput dynamics at the memory cliff.}
The dramatic throughput collapse for BP at batch $\geq\!96$ on VGG11 $224\!\times\!224$ (Sec.~\ref{sec:systems}) is consistent with the GPU driver spilling allocations into shared host memory once device memory is exhausted (the Windows CUDA driver can fall back to system memory instead of raising an out-of-memory error); we did not instrument the driver to confirm this. Latency per step shifts from device-bound ($\sim\!400$ ms at batch $64$) to host-IO-bound ($>\!8$ s at batch $128$). Smaller models or larger device memory (e.g.,\ RTX 4090 $24$\,GB) push the cliff further out and the same-batch advantage shrinks to the modest in-VRAM numbers reported in the main text.

\paragraph{Fair BP baselines.}
We compare DTG-FF at batch $128$ on VGG11 $224\!\times\!224$ against three BP variants designed to recover memory headroom (all measured directly):
\begin{description}
	\tightlist
	\item[BP-VGG11 (vanilla, b=128).] $14$ imgs/s, $8.18$ GB peak (host-memory spill).
	\item[BP+grad-accum (micro-batch $64\!\times\!2$).] Effective batch $128$, $4.18$ GB peak, $157$ imgs/s. Achieves identical effective batch with no device-memory penalty but at the cost of doubling the number of forward/backward passes.
	\item[BP+activation-checkpointing (b=128, every conv block).] $6.35$ GB peak, $92$ imgs/s ($1.5\!\times$ slower than DTG-FF, from recomputation).
\end{description}

\textbf{Honest reading.} Under same effective batch and at the no-spill operating point, BP+grad-accum at $4.18$ GB / $157$ imgs/s dominates DTG-FF ($7.90$ GB / $138$ imgs/s) on both memory and throughput; BP+ckpt at $6.35$ GB / $92$ imgs/s recovers in-VRAM operation at a recompute penalty. DTG-FF beats only vanilla BP at its spill point---a regime practitioners avoid via standard tooling. The structural $\mathcal{O}(L)\!\to\!\mathcal{O}(1)$ activation-memory property of pipelined FF is realized but does not translate into measured systems dominance over memory-optimized BP on this hardware.

\paragraph{Caveats.}
We report wall-clock on a single laptop GPU at modest scale; we do not claim parity with engineered memory-saving stacks like ZeRO/FSDP \citep{rajbhandari2020zero,zhao2023fsdp}. The experiment establishes that the in-principle $\mathcal{O}(L)\to\mathcal{O}(1)$ activation memory property of strict layer-local training is realizable in practice with a trivial code path (single-conv-pass pipelining).

\subsection{Optimizer Protocol Asymmetry Control}\label{app:opt_asymmetry}

DTG-FF uses seven per-layer AdamW optimizers with per-layer cosine schedules and per-layer gradient clipping; the BP-DeepSup baseline in Sec.~\ref{sec:sota} uses a single global AdamW optimizer and global cosine schedule. To rule out the optimizer protocol as a confound, we train BP-DeepSup with the same per-layer protocol (seven per-layer AdamWs matching DTG-FF exactly). The result: \textbf{93.74\%} on CIFAR-10, within $0.01$ pp of the global-optimizer variant (93.73\%). The optimizer protocol is not a meaningful source of the $2.40$ pp FF--BP gap on CIFAR-10. The gap reflects genuine algorithmic differences (detach vs.\ end-to-end gradients, goodness vs.\ learned linear head, random projection vs.\ learned classifier), not a training-setup artifact.

\section{Synthetic Experiment Details}\label{app:synth}

\paragraph{Setup.}
A 3-layer ReLU teacher network with $d_{\mathrm{in}}=50$, $d_{\mathrm{hidden}}=128$, and random output layer of size $K$ (different random draw per seed) generates $(x,y)$ pairs: 20{,}000 train + 5{,}000 test. Inputs are Gaussian $\mathcal N(0,I_d)$; labels are $\arg\max$ of the teacher's logits. Students have 4 hidden layers of size 128 each. All methods share identical data, optimizer family (Adam, $\mathrm{lr}\!=\!10^{-3}$, batch 256), and training budget (8{,}000 steps). Seeds: $\{42, 123, 456, 789, 1024\}$. Class counts: $K\!\in\!\{5,10,15,20,30,50\}$.

\paragraph{Baselines.}
\begin{description}
	\item[Single BP.] One MLP with 4 hidden layers + $K$-way output, end-to-end CE.
	\item[BP-Ensemble.] Four independently initialized single-BP students, softmax-averaged at inference. $\sim\!4\!\times$ the parameter count of DTG-FF.
	\item[BP-DeepSup.] Single MLP with 4 auxiliary linear heads (one per hidden layer). Training loss is summed CE over all heads. Inference is logit-sum over heads. Matched to DTG-FF in backbone, depth, per-layer head count, and inference aggregation; the methods differ in detach boundary, local objective (CE vs.\ goodness), and head parameterization (learned linear vs.\ fixed random projection). See Sec.~\ref{sec:limitations} for the three-dimensional difference.
	\item[DTG-FF.] Per-layer local CE on $\tilde{\mathbf{u}}_l^\top R_l + \mathbf{b}_l$ with spatial goodness $\mathbf{u}_l$ (though for the 1-D synthetic input we use $\mathbf{u}_l = \mathbf{z}_l^2$ directly, a fixed-dimension analog of the CNN spatial goodness), fixed $R_l, \mathbf{b}_l\!\sim\!\mathcal N(0,1/K)$, logit-sum inference, \texttt{detach} between layers.
\end{description}

\paragraph{Parameter counts (per method, $K=10$).}
All four students share the same 4-layer backbone ($d_{\mathrm{in}}\!=\!50$, $d_{\mathrm{hidden}}\!=\!128$); each \texttt{nn.Linear} carries the default learnable bias. DTG-FF's per-layer random readouts $R_l\!\in\!\mathbb R^{128\times 10}$ and $\mathbf b_l\!\in\!\mathbb R^{10}$ are registered as buffers (not learned). Counts measured directly from the implementations:
\begin{itemize}
    \tightlist
    \item \textbf{DTG-FF}: backbone $50\!\cdot\!128 + 128 + 3(128\!\cdot\!128 + 128) = 56{,}064$ plus $4$ learnable temperatures $\alpha_l$ = $\mathbf{56{,}068}$ learnable; fixed buffers $4(128\!\cdot\!10 + 10) = 5{,}160$; total $61{,}228$.
    \item \textbf{BP-DeepSup}: backbone $56{,}064$ plus $4$ learned heads $4(128\!\cdot\!10 + 10) = 5{,}160$ = $\mathbf{61{,}224}$ learnable.
    \item \textbf{Single BP}: $56{,}064 + (128\!\cdot\!10 + 10) = \mathbf{57{,}354}$ learnable.
    \item \textbf{BP-Ensemble}: $4 \!\cdot\! 57{,}354 = \mathbf{229{,}416}$ learnable ($\approx 4\!\times$ DTG-FF).
\end{itemize}
The DTG-FF vs.\ BP-DeepSup learnable-parameter mismatch is $5{,}156$ params (the four per-layer heads, less the four $\alpha_l$ scalars), about $9\%$ of backbone size, in BP-DeepSup's favor. We retain this mismatch rather than tying $R_l$ in DTG-FF because doing so would change the algorithm; it should be noted when interpreting capacity-matched comparisons. The ratio is stable across $K$.

\paragraph{Paired-difference methodology.}
The random teacher varies by seed, so absolute accuracies have high variance (std $4$--$17\%$ across seeds). We control this by reporting per-seed paired differences (DTG-FF minus baseline) in Table~\ref{tab:synthetic}; this cancels the teacher-induced shared variance.

\begin{table}[h]
	\centering
	\small
	\setlength{\tabcolsep}{4pt}
	\renewcommand{\arraystretch}{0.95}
	\caption{Paired differences (DTG-FF minus baseline accuracy, per seed) on synthetic teacher--student tasks, 5 seeds. $(n/5)$ shows the number of seeds where DTG-FF wins. Summary in Sec.~\ref{sec:synthetic}.}\label{tab:synthetic}
	\begin{tabular}{@{}rccc@{}}
		\toprule
		$K$ & DTG-FF $-$ single BP     & DTG-FF $-$ BP-DeepSup    & DTG-FF $-$ BP-Ensemble   \\
		\midrule
		5   & $+0.38\!\pm\!0.84$ (3/5) & $-0.23\!\pm\!0.58$ (2/5) & $-1.74\!\pm\!1.26$ (0/5) \\
		10  & $+1.20\!\pm\!1.26$ (4/5) & $+0.84\!\pm\!0.99$ (3/5) & $-1.60\!\pm\!1.09$ (0/5) \\
		15  & $+1.44\!\pm\!0.60$ (5/5) & $+0.48\!\pm\!0.62$ (3/5) & $-2.06\!\pm\!0.28$ (0/5) \\
		20  & $+1.32\!\pm\!1.05$ (5/5) & $+0.55\!\pm\!1.20$ (3/5) & $-1.86\!\pm\!1.16$ (0/5) \\
		30  & $+3.39\!\pm\!2.36$ (5/5) & $+1.57\!\pm\!1.45$ (4/5) & $-1.54\!\pm\!1.11$ (0/5) \\
		50  & $+2.63\!\pm\!1.27$ (5/5) & $+2.00\!\pm\!0.95$ (5/5) & $-0.75\!\pm\!0.51$ (0/5) \\
		\bottomrule
	\end{tabular}
\end{table}

\begin{figure}[t]
	\centering
	\includegraphics[width=0.78\textwidth]{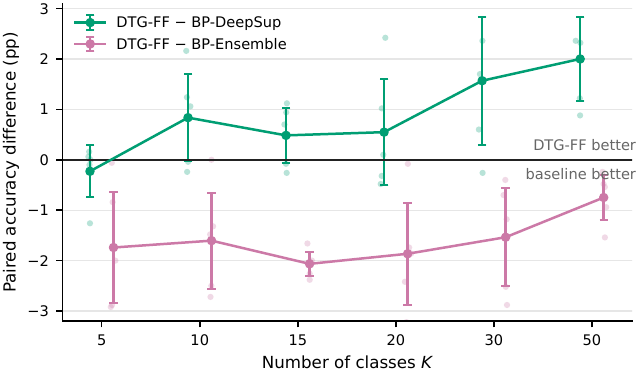}
	\caption{Architecture-matched synthetic controls. Points show seed-level paired differences; markers and error bars show mean $\pm$ 95\% normal-approximation CI across five seeds. DTG-FF becomes favorable relative to BP-DeepSup at larger $K$ in this controlled teacher--student setting, but remains below the $4{\times}$-parameter BP ensemble at every $K$.}
	\label{fig:bpcontrols}
\end{figure}

\paragraph{Pre-specified trend test.}
Following the hypothesis ``DTG-FF's advantage over BP-DeepSup emerges more clearly at larger $K$,'' we partition $K\!\in\!\{5,10,15,20\}$ (low-$K$, $n\!=\!20$ seed-$K$ observations) vs.\ $K\!\in\!\{30,50\}$ (high-$K$, $n\!=\!10$) and compute mean paired DTG-FF $-$ BP-DeepSup. Low-$K$ mean: $+0.41\%$. High-$K$ mean: $+1.78\%$. Difference of means: $+1.37$ pp. Bootstrap 95\% CI (paired observations resampled with replacement, $n_{\mathrm{boot}}=10{,}000$, seed $42$): $[+0.59, +2.15]$. The CI does not cross zero, supporting the regime-qualified claim.

\paragraph{Train--test gap analysis.}
To assess whether DTG-FF's advantage reflects implicit regularization rather than a purely representational effect, we record training accuracy alongside test accuracy. Table~\ref{tab:traintest} reports per-method train--test gaps averaged over 5 seeds. BP and BP-DeepSup overfit to near-100\% training accuracy at all $K\!\geq\!10$, producing train--test gaps of $21$--$34$ percentage points. DTG-FF reaches only $86$--$97\%$ on training at $K\!\geq\!10$, with train--test gaps of $15$--$18$ pp. The gap \emph{difference} between BP and DTG-FF is generally larger at higher $K$ ($1.4$ pp at $K\!=\!5$; $16.7$ pp at $K\!=\!30$). This is consistent with local per-layer training acting as implicit regularization---each layer cannot freely coordinate with others to memorize the training set---and partially explains why DTG-FF can reach comparable or higher test accuracy despite fitting the training set less tightly than end-to-end BP variants.

\begin{table}[t]
	\centering
	\small
	\setlength{\tabcolsep}{4pt}
	\renewcommand{\arraystretch}{0.95}
	\caption{Test accuracy and train--test gap ($\Delta$ = train acc $-$ test acc, both in \%) on synthetic teacher--student tasks, 5 seeds. Last column: how much smaller the DTG-FF gap is than the BP gap. DTG-FF overfits less at every $K$; the effect is generally larger at higher $K$.}\label{tab:traintest}
	\begin{tabular}{@{}cccccc@{}}
		\toprule
		$K$ & BP (test / $\Delta$) & BP-Ens (test / $\Delta$) & BP-DS (test / $\Delta$) & DTG-FF (test / $\Delta$) & BP $\Delta$ $-$ DTG $\Delta$ \\
		\midrule
		5   & 83.8 / +15.9         & 85.9 / +14.1             & 84.4 / +15.0            & 84.2 / +14.5             & +1.4                         \\
		10  & 77.6 / +22.3         & 80.4 / +19.6             & 78.0 / +21.5            & 78.8 / +17.9             & +4.4                         \\
		15  & 76.0 / +23.9         & 79.5 / +20.5             & 77.0 / +22.4            & 77.5 / +18.4             & +5.5                         \\
		20  & 77.6 / +22.2         & 80.8 / +19.2             & 78.4 / +20.7            & 79.0 / +15.0             & +7.2                         \\
		30  & 65.2 / +34.2         & 70.1 / +29.9             & 67.0 / +30.5            & 68.6 / +17.5             & \textbf{+16.7}               \\
		50  & 72.9 / +27.1         & 76.3 / +23.7             & 73.6 / +25.7            & 75.6 / +15.6             & +11.5                        \\
		\bottomrule
	\end{tabular}
\end{table}

\end{document}